\documentclass[letterpaper]{article} 
\usepackage[]{aaai23}  
\usepackage{times}  
\usepackage{helvet}  
\usepackage{courier}  
\usepackage[hyphens]{url}  
\usepackage{graphicx} 
\usepackage{natbib}  
\usepackage{caption} 
\usepackage{algorithm}
\usepackage{algorithmic}

\usepackage{newfloat}
\usepackage{listings}
\DeclareCaptionStyle{ruled}{labelfont=normalfont,labelsep=colon,strut=off} 
\floatstyle{ruled}
\newfloat{listing}{tb}{lst}{}
\floatname{listing}{Listing}
\graphicspath{{./figs/}}
\usepackage{subcaption}
\usepackage{mathtools,xparse}
\DeclarePairedDelimiter{\norm}{\lVert}{\rVert}
\usepackage{amsmath,amssymb} 
\usepackage{makecell}
\usepackage{cite}
\usepackage{hyperref}
\hypersetup{
    colorlinks=true,
    linkcolor=cyan,
    filecolor=magenta,      
    urlcolor=magenta,
    citecolor=black
    }
\title{ProCST: Boosting Semantic Segmentation Using Progressive Cyclic Style-Transfer}
\author {
    Shahaf Ettedgui,
    Shady Abu-Hussein,
    Raja Giryes 
}
\affiliations {
    Tel Aviv University\\
}

\begin{document}

\maketitle

\begin{abstract}
Using synthetic data for training neural networks that achieve good performance on real-world data is an important task as it can reduce the need for costly data annotation. Yet, synthetic and real world data have a domain gap. Reducing this gap, also known as domain adaptation, has been widely studied in recent years.
Closing the domain gap between the source (synthetic) and target (real) data by directly performing the adaptation between the two is challenging. In this work, we propose a novel two-stage framework for improving domain adaptation techniques on image data. In the first stage, we progressively train a multi-scale neural network to perform image translation from the source domain to the target domain.
We denote the new transformed data as ``Source in Target'' (SiT). Then, we insert the generated SiT data as the input to any standard UDA approach. This new data has a reduced domain gap from the desired target domain, which facilitates the applied UDA approach to close the gap further. We emphasize the effectiveness of our method via a comparison to other leading UDA and image-to-image translation techniques when used as SiT generators. Moreover, we demonstrate the improvement of our framework with three state-of-the-art UDA methods for semantic segmentation, HRDA, DAFormer and ProDA, on two UDA tasks, GTA5 to Cityscapes and Synthia to Cityscapes.
Code and translated datasets available at \url{https://github.com/shahaf1313/ProCST}.
\end{abstract}

\begin{figure}[t!]
    \centering
    \includegraphics[width=\linewidth,height=8cm,keepaspectratio]{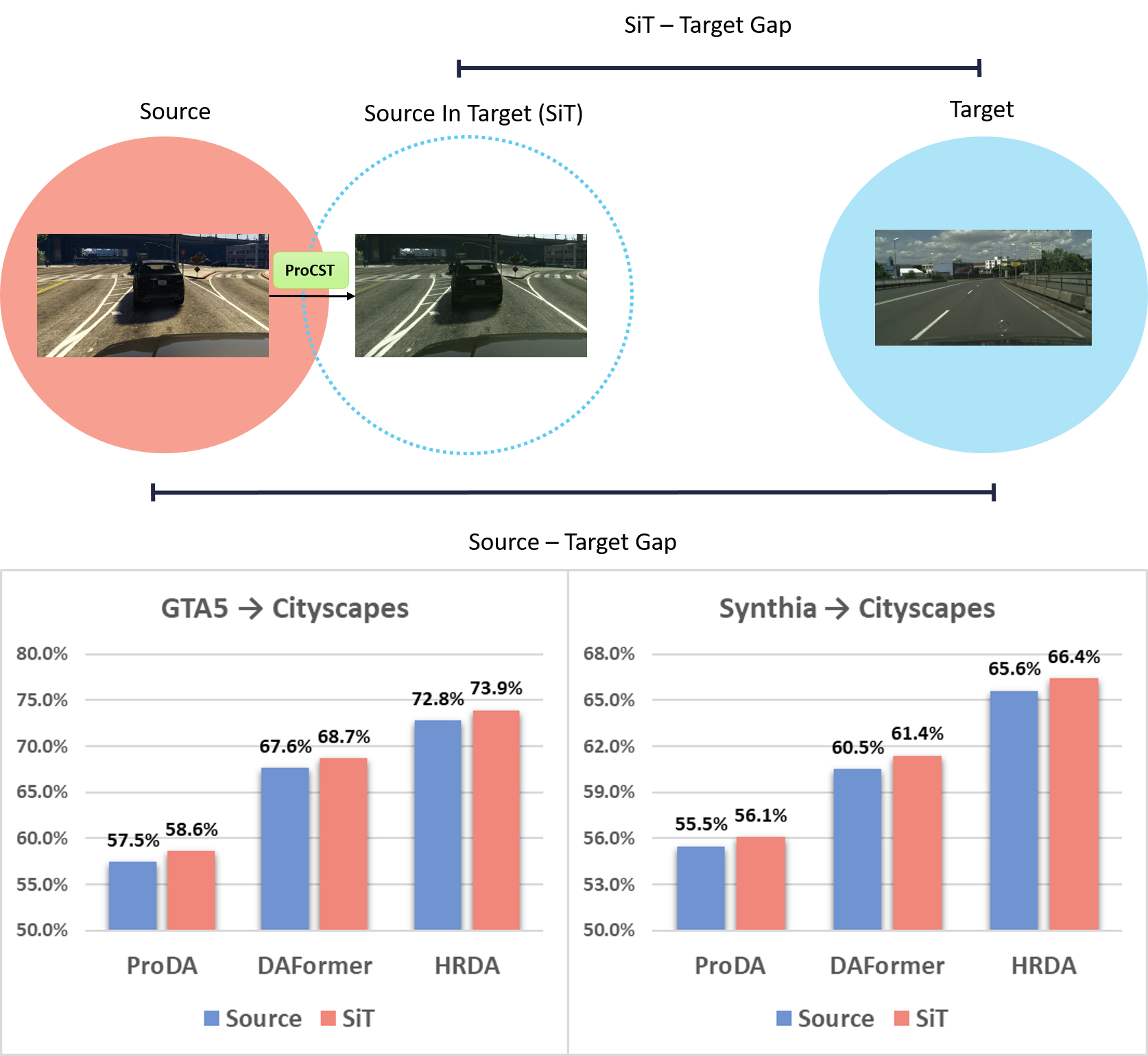}
    \caption{{\bf Domain Gap Reduction using SiT images.} Images from the source domain are translated onto a SiT domain, using ProCST. The translated image preserves the content of the source image, but changes the appearance and style to the ones of the target domain for reducing the domain gap. This helps existing UDA methods to learn improved semantic segmentation models for the target domain. The bottom chart shows the achieved improvement in semantic segmentation (in terms of mIoU) when we use SiT data for training state-of-the-art UDA methods.}
    \label{fig:procst_teaser}
\end{figure}

\begin{figure*}
    \centering
    \includegraphics[width=\textwidth]{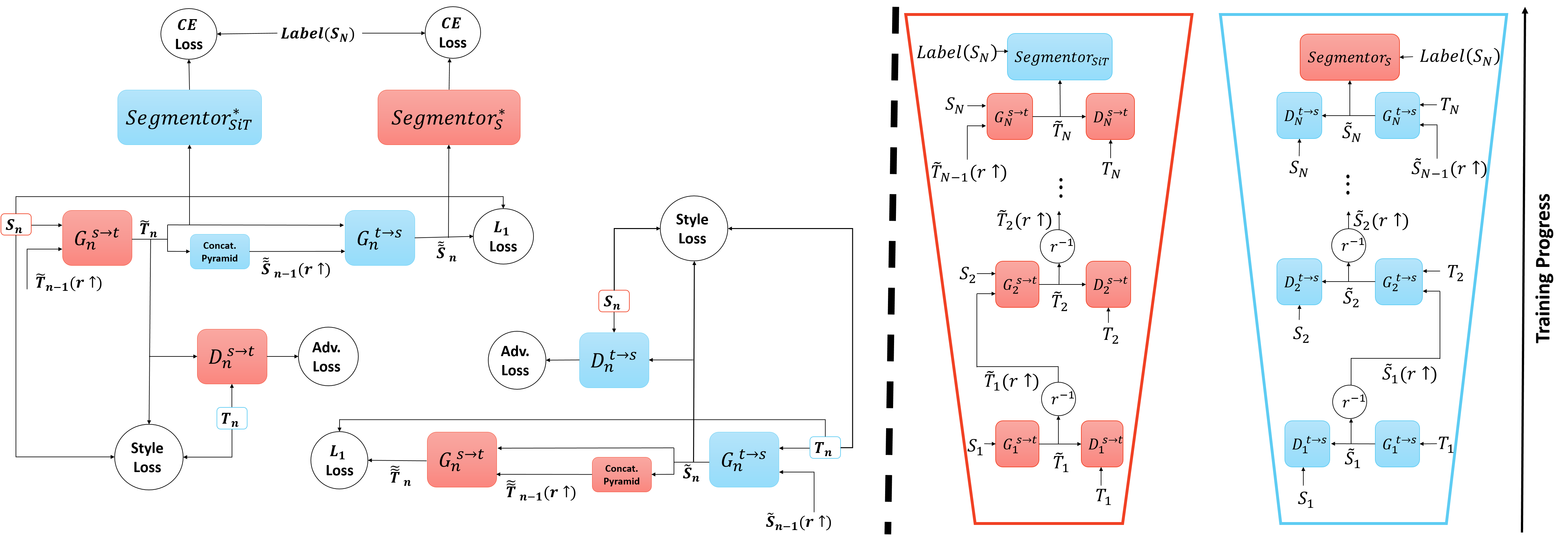}
    \caption{{\bf ProCST model.} Networks are colored by their input domain: red for source domain and blue for target domain. On the right, the bi-directional multiscale image pyramid structure that perform  $\mathbb{S}\xrightarrow{}\mathbb{T}$ and $\mathbb{T}\xrightarrow{}\mathbb{S}$ image translation, respectively. On the left, a flowchart of our used losses  for all the pyramid levels. The highest level $n=N$ has an additional label loss.}
    \label{fig:full_model}
\end{figure*}

\section{Introduction}

Many deep learning approaches have been proposed in the past years for solving the semantic segmentation task, where the goal is to output a segmentation map: a per-pixel label for a given input image. However, as in many cases, a large amount of data is required to train the segmentation networks for achieving a good generalization. In contrast to classification, where a single label is required per each input image, in semantic segmentation each pixel should be labeled, which makes the annotation very time-consuming and prone to inaccuracies due to human error.

Unsupervised Domain Adaptation (UDA) aims at mitigating this lack of annotated training data. It uses the vast amount of annotated data in a given source domain to learn to perform segmentation in a given target domain that lacks annotations.
A popular source dataset is synthetic simulation data that have the per-pixel annotations `for free'. The target is real data with the same classes as in the simulation.

The naive approach of training a segmentation network only using annotated source data fails to achieve good results because of the major gap between real-world and synthetic simulation images. Moreover, training a segmentation network on source images that were translated to the target domain using an image-to-image technique fails to achieve good results on the target domain \cite{hoffman2018cycada}. 

Many UDA techniques have been proposed to achieve good performance in the segmentation of the target domain. 
In these approaches, the training of the segmentation neural network incorporates real images, which do not have ground-truth labels, together with the synthetic images and their annotations.
For example, some suggested training a neural network using the annotated samples from the source domain, and then either adapt the network to the target domain \cite{zhang2021prototypical} or, alternatively, transfer the samples of the source domain to the target while preserving the structure of the source images, such that the segmentation maps of the source can be used for the transferred images as labels \cite{wang2021consistency, li2019bidirectional, yang2020fda}. 

These and most of the current methods try to close the entire domain gap at once. In this work, we offer a prior step to these methods, whose goal is to reduce the gap as a first step. This idea is depicted in Figure~\ref{fig:procst_teaser}. We suggest a Progressive Cyclic Style-Transfer (ProCST) network, which has a multiscale architecture that uses progressive training for performing style transfer. This model uses the annotations of the source domain to preserve the semantic information in the transfer. Figure~\ref{fig:full_model} presents the ProCST architecture. 

ProCST narrows the domain gap between the source and target domains, by translating the source images to mimic images from the target domain, which we refer to as source-in-target (SiT). The SiT dataset is closer to the target domain and preserves the images content. It can be used as input to any UDA method and improve it as we show hereafter.

We also show that simply taking an off-the-shelf approach to generate such a SiT dataset fails to improve UDA, thus highlighting the importance of our proposed model and loss functions for generating SiT.
The experiments we conduct indicate that existing UDA and image-to-image translation methods fail to preform well as SiT data generators. Those methods tend to either distort or delete objects from the source image, and thus the translated images do not contain content equal to the original image, as can be seen for example in Figure~\ref{fig:epe_compare}. Thus, the original source annotations are no longer accurate and are only partially valid. Nevertheless, ProCST preserves the content of the source image due to the unique combination of losses that we suggest for the training process, with our novel cyclic label loss among them. Hence, we can use the SiT data instead of the source domain data as input to state-of-the-art UDA methods for semantic segmentation. We show that the new SiT data improves their performance. 

We evaluate ProCST on UDA for semantic segmentation, where the target domain in both cases is Cityscapes \cite{cityscapes}, and the source domains are either GTA5 \cite{GTA5} or Synthia \cite{synthia}. We demonstrate the effectiveness of our scheme using three state-of-the-art methods, namely, ProDA \cite{zhang2021prototypical}, DAFormer \cite{hoyer2021daformer} and HRDA \cite{hoyer2022hrda}. We measure the Mean Intersection Over Union (mIoU) over the Cityscapes test data, and improve the performance of all these approaches. For example, we improve over HDRA, which is the current state-of-the-art, by 1.1\% mIoU on GTA5 dataset. This demonstrates the effectiveness of our proposed framework.

\section{Related Work}

\textit{Semantic Segmentation} has been examined widely using deep learning methods. \citet{long2015fully, yu2015multi, chen2017deeplab, zhao2017pyramid} propose various architectures and train them on pixel-wise manually annotated datasets. However, the annotation procedure is time-consuming. Training using small datasets usually leads to poor segmentation quality. One way to overcome availability of annotations is to use synthetic data, where annotations are available generously. Yet, due to the domain gap between the synthetic and the real data, the resulting model does not generalize well on the latter.

\textit{Unsupervised Domain Adaptation} methods \cite{hoffman2018cycada,wang2018deep, gao2021dsp, zhang2019category,Wilson2020UDA}, seek to align domains' distributions. \citet{zellinger2017central, geng2011daml, mancini2018boosting} adapt the source domain to the target domain by minimizing discrepancy measures and matching the statistical moments of the two domains. Yet, these methods struggle to align the domains appropriately, which leads to poor generalization over the target domain.

Another popular strategy in domain adaptation is \textit{Adversarial Learning}. \citet{ganin2015unsupervised, tzeng2017adversarial, shu2018dirt, kumar2018co} use a discriminator to distinguish between target domain images and translated source domain images, where the translator (generator) tries to fool the discriminator. As a result, the generator learns to map the source images to the target domain. \citet{richter2021enhancing} utilize buffers from the simulator used to create the synthetic images for guiding the translation network to generate better looking images. This technique outputs good image quality but it fails to preserve image content, and thus preforms poorly when incorporated in a semantic segmentation framework as shown in Sec.~\ref{sec:experiment}.  

For domain adaptation, a successful training strategy is the addition of cyclic loss to the total objective. \citet{hoffman2018cycada, zhu2017unpaired, yi2017dualgan} translated source images to the target domain and then translated back to the source domain and required consistency. Note that when used for UDA, the cyclic loss is applied with other loss functions as translating the data alone without using the segmentation map information results in poor performance.

One major step that enabled high-resolution image generation was progressive training \citet{karras2017progressive}. The proposed training procedure first trains the network to generate low-resolution images, then adds layers and increases the generated images size sequentially. This concept enabled the generation of high resolution images with superior quality and was employed in many works \cite{karras2017progressive, karras2019style, karras2020training, karras2021alias, wang2018high, shaham2019singan}.

\begin{figure*}[t]
    \centering
    \begin{subfigure}{0.24\linewidth}
        \centering
        \includegraphics[width=\linewidth]{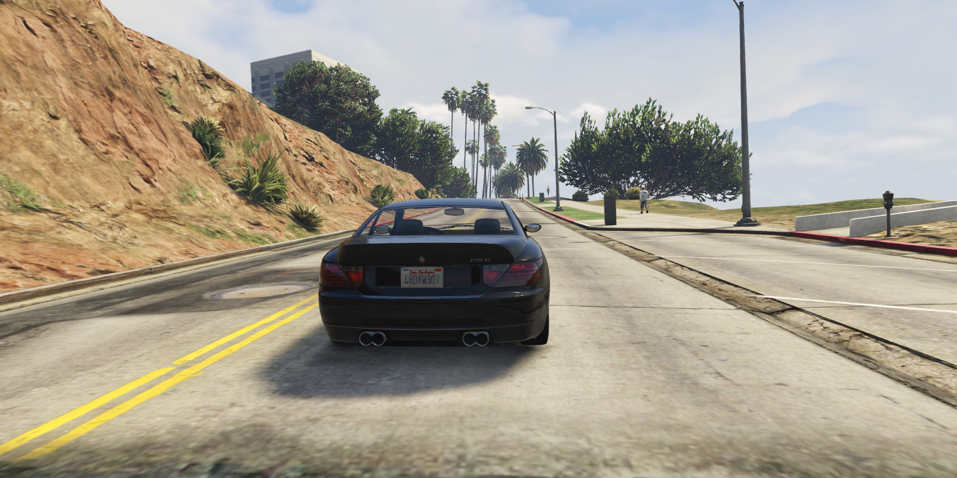}
    \end{subfigure}
    \begin{subfigure}{0.24\linewidth}
        \centering
        \includegraphics[width=\linewidth]{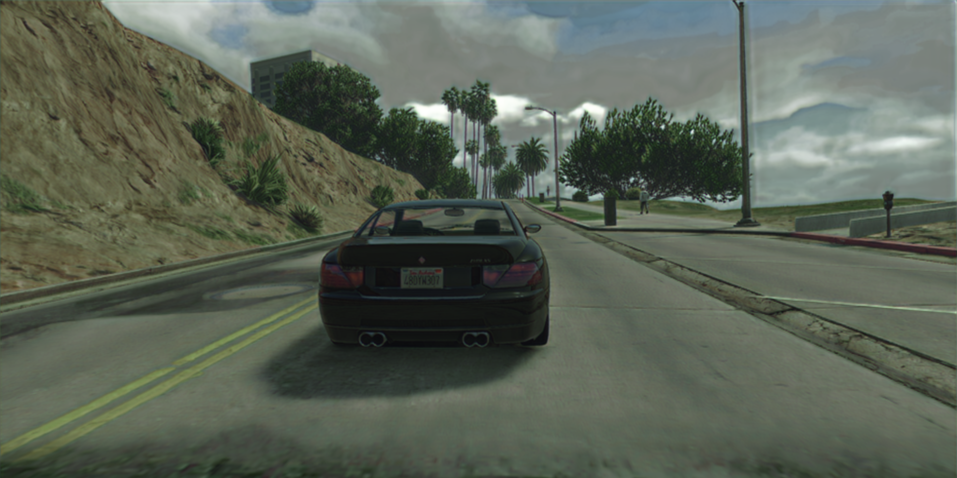}
    \end{subfigure}
    \begin{subfigure}{0.24\linewidth}
        \centering
        \includegraphics[width=\linewidth]{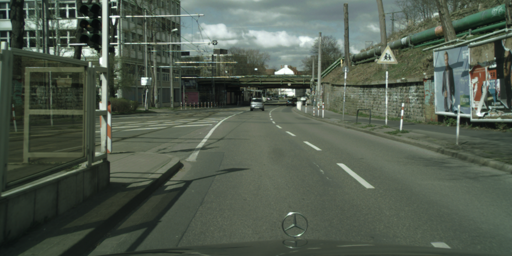}
    \end{subfigure}
    \begin{subfigure}{0.24\linewidth}
        \centering
        \includegraphics[width=\linewidth]{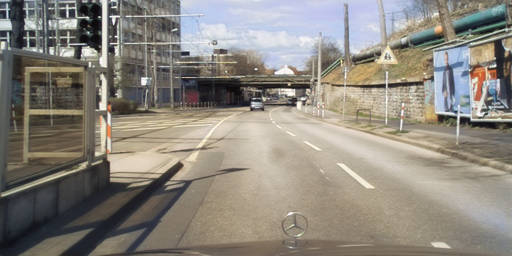}
    \end{subfigure}
    
    \centering
    \begin{subfigure}{0.24\linewidth}
        \centering
        \includegraphics[width=\linewidth]{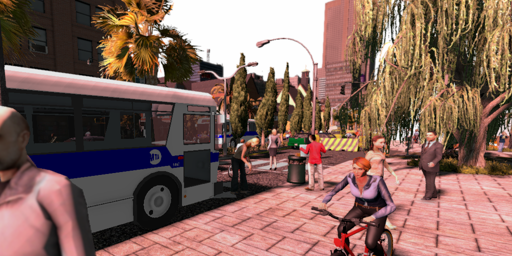}
    \end{subfigure}
    \begin{subfigure}{0.24\linewidth}
        \centering
        \includegraphics[width=\linewidth]{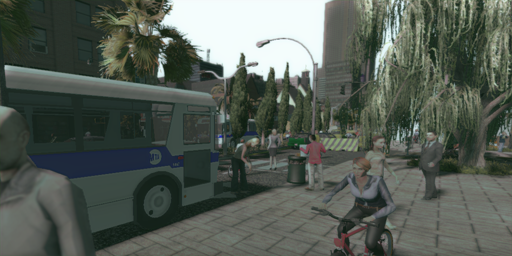}
    \end{subfigure}
    \begin{subfigure}{0.24\linewidth}
        \centering
        \includegraphics[width=\linewidth]{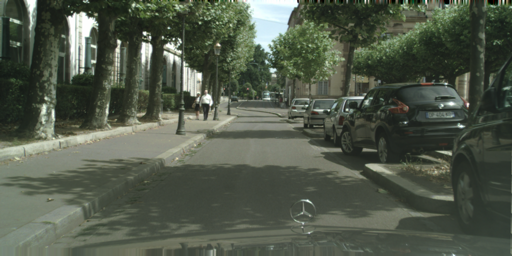}
    \end{subfigure}
    \begin{subfigure}{0.24\linewidth}
        \centering
        \includegraphics[width=\linewidth]{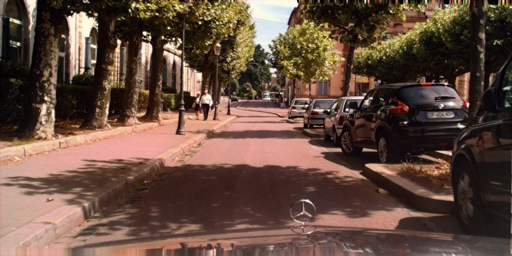}
    \end{subfigure}

    \caption{{\bf Bi-directional translation of ProCST.} Top row: $\text{GTA5} \xleftrightarrow[]{} \text{Cityscapes}$ image translation; Bottom row: $\text{Synthia} \xleftrightarrow[]{} \text{Cityscapes}$ image translation. Both rows, left to right: source, SiT, target, and ``Target in Source'' translated images.}
    \label{fig:sit_tis_translations}
\end{figure*}

\citet{zhang2021prototypical, hoyer2021daformer} decrease the domain gap using self-supervised learning (SSL). For instance, \citet{zhang2021prototypical} use pseudo label denoising and target structure learning in a three stage training process. \citet{hoyer2021daformer} base their network architecture on two vision transformers \cite{liang2021swinir, dosovitskiy2020image}, in a student-teacher training framework fed with domain-mixed augmented images. \citet{hoyer2022hrda} extend this concept further, and combine multi-resolution training that fuses low-resolution context with high resolution crop using an attention mechanism. 

In this work, we propose an algorithm for generating ``Source in Target'' (SiT) images, namely, a domain shifted version of the original source images, which narrows the source-target domain gap. We then show that the current leading UDA methods perform better when trained using our SiT images instead of the original source images.


\section{Method}
\label{sec:method}

We present ProCST, an image-to-image translation framework that is used as a zero-step before applying existing UDA methods for improving their performance. ProCST is a multiscale network trained with a cyclic style transfer loss using a progressive training scheme. 
ProCST also uses two semantic segmentation networks that utilize source domain labels in the training procedure. 

Our multiscale training is based on an image pyramid structure: a list of generators that creates a full scale translated image, where each generator is responsible for translating the image at a certain scale. The generator has two inputs: the first is a translated image from the generator of the previous scale, upscaled to the current scale size. The second is a resized version of the original source image that matches the current generator scale. Note that the generator of the lowest scale receives only the second input. The output of each generator is a translated image in the scale it is using. This process is depicted in Figure~\ref{fig:full_model}.

Each scale consists of two generators: $G^{st}$ for translating images from the source domain to the target domain, and $G^{ts}$ for the inverse translation. Each of the generators has a separate discriminator, $D^{st}$ and $D^{ts}$, respectively. Both the generators and the discriminators are CNNs with depth that increases with the resolution. A similar approach has been used to train a GAN using a single image for the task of image generation that receives noise and outputs an image \cite{shaham2019singan}, which is very different from the translation task of our method that transfers an image from one domain to another.  

The training process is progressive: we train scales sequentially beginning from the lowest. This training approach has proven to create realistic and clean images, as shown for the Progressive GAN \cite{karras2017progressive}, which is used for image generation (and not translation as we do here). 

The losses are a unique combination of the adversarial-cyclic-style criteria, which allow changing the visual style of the image but also preserving the original content. In addition, when training the full resolution image generators, we incorporate a label loss term and present the novel cyclic label loss. These terms are described in Section~\ref{sec:losses}. We start with basic definitions that are used to describe our model.

\paragraph{Model definitions}
Let $\mathbb{S}$ be the source domain and $\mathcal{S}$ the source dataset with images ${s}^{(l)}$ and annotations ${Label(s^{(l)})}$, 
\begin{equation}
\begin{aligned}
    &\mathbb{S} \subset \mathbb{R}^{H_S \times W_S \times 3} \times \mathbb{R}^{H_S \times W_S \times K} 
    \\
    &{\mathcal{S}}=\{({s}^{(l)}, {Label(s^{(l)})}) \in \mathbb{S}\}_{l=0}^{N_S},
\end{aligned}
\end{equation}
where $H_S, W_S$ are the height and width of source domain images, $K$ is the number of class categories and $N_S$ is the number of data points in the source dataset. Also, let $\mathbb{T}$ be the target domain, and $\mathcal{T}$ the target dataset with images ${t}^{(l)}$,
\begin{equation}
\begin{aligned}
    &\mathbb{T} \subset \mathbb{R}^{H_T \times W_T \times 3}
    \\
    &{\mathcal{T}}=\{{t}^{(l)} \in \mathbb{T}\}_{l=0}^{N_T},
\end{aligned}
\end{equation}
where $H_T, W_T$ are the height and width of target domain images, and $N_T$ is the number of data points in the target dataset.
Notice that the target domain annotations are not available, and we assume that source and target datasets share the same class categories, $c=1,..,K$.
Denote $N$ the number of scales in the multiscale model of size $N$,
and denote by $0<r<1$ the scale factor between two adjacent scales.
Then, at scale $n=1,..,N$, the dimensions are $(H_n,W_n)= (r^{N-n} H_N, r^{N-n} W_N)$, where $H_n$ and $W_n$ are the height and width of the images at scale $n$.

 \paragraph{Model architecture}
ProCST is a multiscale model. Scales are trained sequentially in a progressive training process. Each scale has two generators, $G_n^{st}$ and $G_n^{ts}$ and two discriminators, $D_n^{st}$ and $D_n^{ts}$, where $n$ denotes the scale level. 

We turn to describe the $\mathbb{S}\rightarrow\mathbb{T}$ translation process, where the complementary translation $\mathbb{T}\rightarrow\mathbb{S}$ is symmetric. 
%
At the $n$th scale, $n=1,..,N$, ProCST generates SiT fake images using the generator chain $G_k^{st}, k\le n$. Each generator receives the upscaled output of the previous generator and the current source image resized to the current scale's size. This process is repeated till we reach the last scale. More formally, let $s \in \mathcal{S}$ be a source image, and  $t \in \mathcal{T}$ be a target image. Let $s_n, t_n$ be the resized source and target images to scale $n$, respectively. A fake SiT image $\Tilde{t}_{n}$ is generated using $n$ generators $G_k^{st}, \ k \le n$ by the following procedure: 
\begin{equation}
    \begin{cases}
      \Tilde{t}_{n} = G_{n}^{st}[s_{n},\Tilde{t}_{n-1}(r\uparrow)] & n>1\\
      \Tilde{t}_{1} = G_{1}^{st}[s_{1}] & n=1,
    \end{cases}
\end{equation}
where $(r\uparrow)$ means the image is upscaled by a factor of $r^{-1}$. 
For instance, on the right-hand-side of Figure~\ref{fig:full_model} we can see that the second scale receives an upscaled output image from the first generator together with the resized source image. Both images share the same size: the size of the second scale. We use the above translation process to generate all the required images to obtain the unique combination of loss terms, which varies as a function of $n$. This loss combination, described deeply in Section~\ref{sec:losses}, is the heart of ProCST. 


\paragraph{Boosting generic UDA methods}
We turn to describe how ProCST may be used to improve existing UDA methods. 
These methods receive as input a source dataset and a target dataset, and output segmentation maps of the target dataset:
\begin{equation}
    \text{UDA}[\mathcal{S}, \mathcal{T}] \xrightarrow{} Label(\mathcal{T}).
\end{equation}
These methods encounter a large gap between the source and target domains.
Training a ProCST model results in a chain of domain adapters, $\{G_n^{st},G_n^{ts}\}_{n=1}^N$, that reduces this gap. Let $\text{ProCST}_{\mathbb{S}\xrightarrow{}\mathbb{T}}$ be a trained ProCST model that translates images from a source domain $\mathbb{S}$ to the target domain $\mathbb{T}$. We can build a new SiT  dataset using this model, with the same annotation maps as the original source dataset:
\begin{equation}
    \begin{aligned}
        \text{SiT} &= \{(\text{ProCST}_{\mathbb{S}\xrightarrow{}\mathbb{T}}\{s\}, Label(s)), \forall s \in \mathcal{S}\}, \\
        & = \{(\Tilde{t}^{(l)}, Label(s^{(l)}))\}_{l=1}^{N_S}.
    \end{aligned}
\end{equation}
The $\text{SiT}$ is an in-between dataset, i.e., its domain gap from the desired target dataset is smaller than the original source dataset - as depicted in Figure~\ref{fig:procst_teaser}.
One may now train any arbitrary UDA method with the domain adapted $\text{SiT}$ data:
\begin{equation}
    \text{UDA}[\text{SiT}, \mathcal{T}] \xrightarrow{} Label(\mathcal{T}).
\end{equation}
In this way, the same UDA method encounters a smaller domain gap and thus achieves increased accuracy. This is a generic approach that boosts performance of UDA methods by narrowing the domain gap for them.

\section{ProCST Loss Combination for UDA}
\label{sec:losses}
For transforming the source data into SiT data in our framework, we propose the following novel combination of loss functions that adapt the style of the target while keeping the content of the source. This combination is specifically tailored for the task of UDA to make sure that the segmentation performance are improved when training UDA methods using the translated SiT data. We start by describing the cyclic style transfer loss.
Without loss of generality, we will focus on the $\mathbb{S}\xrightarrow{}\mathbb{T}$ translation. The inverse direction is symmetric for this loss bundle. 
Recall that $s_n, \Tilde{t}_n,$ and $t_n$ have already been defined earlier in Section~\ref{sec:method}.  

\paragraph{Adversarial loss} We train a full CNN discriminator for each scale, $D_n^{st}$. We use the Wasserstein GAN loss with gradient penalty \cite{gulrajani2017improved}. $D_n^{st}$ receives $(\Tilde{t}_n, t_n)$ as fake and real images, respectively. The loss is given by 
\begin{equation}
    \mathcal{L}^{st}_{adv}=\mathcal{L}^{st}_{adv}(\Tilde{t}_n, t_n).
\end{equation}

\paragraph{Style loss} Our method focuses on generating an image taken from a source domain to have the content of the original image but the appearance and style of the target domain. To fulfill this task we incorporate the style transfer loss, shown in \citet{gatys2016image} and \citet{johnson2016perceptual}. For this loss, they use features of a pretrained VGG-19 network. In \citet{gatys2016image}, they have shown that some features control the texture and style and some control the visual content. The offered loss that calculates two matrices:

\textit{Content features} - $\ell_2$ norm between the content image's features and the fake image's features. In our work, $s_n$ is the content image and $\Tilde{t}_n$ is the fake image. 

\textit{Style features} - $\ell_2$ norm between a Gram matrix (correlation matrix) of the style image's features and a Gram matrix of the fake image's features. In our work, $t_n$ is the style image and $\Tilde{t}_n$ is the fake image. Following \citet{johnson2016perceptual}, a total variation regularization is also added to the loss. Thus, a Style Loss term is given by:
\begin{eqnarray}
        && \mathcal{L}^{st}_{style}= 
        \lambda_{content}\mathcal{L}^{st}_{content}(s_n, \Tilde{t}_n) \\ \nonumber
        && ~~~+ \lambda_{style,Gram}\mathcal{L}^{st}_{style,Gram}(t_n, \Tilde{t}_n) 
        + \lambda_{TV}\mathcal{L}^{st}_{TV}(\Tilde{t}_n).
\end{eqnarray}

\paragraph{Cyclic loss} Cyclic training \cite{zhu2017unpaired} is a well known strategy in domain adaptation \cite{hoffman2018cycada}. The main property of the cyclic training regime is the ability to trace back any translated image to the original domain. This is done by calculating the $\ell_1$ loss between the original source image and the same source image that is translated twice (using $G_n^{st}$ followed by $G_n^{ts}$), namely $\norm{s_n-\Tilde{\Tilde{s}}_{n}}_1$, where $\Tilde{\Tilde{s}}_{n}$ denotes the source image that is translated twice.

The cyclic transfer procedure in ProCST is not trivial. First, we generate $\Tilde{t}_n$ as described earlier. Then, we use the generators chain $G_k^{ts}, k\le n$ with a new input image $\Tilde{t}_n$, in order to output the result of the cycle $\Tilde{\Tilde{s}}_n$. Namely, we use
\begin{equation}
    \begin{aligned}
        & \Tilde{\Tilde{s}}_n = G_{n}^{ts}[\Tilde{t}_{n},\Tilde{\Tilde{s}}_{n-1}(r\uparrow)], \\
        & \mathcal{L}^{ss}_{cyclic}=\norm{s_n-\Tilde{\Tilde{s}}_n}_1.
    \end{aligned}
\end{equation} 

\paragraph{Cyclic style transfer (CST) loss}
The above loss terms are unified into one CST loss term, which is evaluated only at the current scale $n$ in a progressive manner. The loss is
\begin{align}
        \nonumber \mathcal{L}_{CST} &=  \lambda_{adv.}\mathcal{L}_{adv.} + \lambda_{style}\mathcal{L}_{style} + \lambda_{cyclic}\mathcal{L}_{cyclic}  \\ \nonumber
         & =\lambda_{adv.}(\mathcal{L}^{st}_{adv.} 
         + \mathcal{L}^{ts}_{adv.}) 
         + \lambda_{style}(\mathcal{L}^{st}_{style} + \mathcal{L}^{ts}_{style}) \\ 
         & + \lambda_{cyclic}(\mathcal{L}^{ss}_{cyclic}+\mathcal{L}^{tt}_{cyclic}).
\end{align}

\paragraph{Label loss}
High resolution images contain a vast amount of knowledge and details. Thus, generating good translation requires fine tuning and as much information as possible. Adding a segmentation loss to image generation networks has proven to be effective \cite{richter2021enhancing}.
Therefore, we use the segmentation maps of the source dataset in order to better adapt to the new domain when more details become present. Specifically, at the last scale of ProCST, we also use a label loss in addition to CST loss. This is the only loss in our model that is not symmetric, i.e., it is not used in the $\mathbb{T}\rightarrow\mathbb{S}$ direction. 

The label loss has two different parts. The first consists of a segmentation network that is trained to segment SiT images, $Segmentor_{SiT}$. The input to this network is a SiT image, and it is trained to output the segmentation map of the original source image. The second part is a novel cyclic label loss. We pretrained a segmentation network on the source dataset, as source annotations are available. Denote this pretrained network as $Segmentor_S$. We take the source image after it has been transferred using the cyclic operation from source to target and then back to source, and enter it to the pretrained $Segmentor_S$ together with the original source segmentation map. This process results in a segmentation loss transfer between $G^{st}$ and $G^{ts}$, which further encourages good segmentation features in both generators.

For both $Segmentor_{SiT}$ and $Segmentor_{S}$, we use ResNet101 \cite{he2016deep} as the segmentation network, which outputs a probability map $p^{i,j,k}$ per class $k$.
Each Segmentor network is trained with the Cross Entropy (CE) loss that is calculated between the segmentor's output $p^{i,j,k} \in \mathbb{R} ^{H \times W \times K} $ and the matching Ground Truth (GT) segmentation map $m^{i,j,k} \in \mathbb{R} ^{H\times W \times K} $. To train the segmentation networks we use the following loss
\begin{equation}
    \mathcal{L}_{CE}[p,m] = -\sum_{i=1}^{H}\sum_{j=1}^{W}\sum_{k=1}^{K}{m^{i,j,k} \log {p^{i,j,k}}}.
\end{equation}

\textit{SiT label loss.} We train $Segmentor_{SiT}$ together with ProCST. It receives as input the translated SiT images and is trained to output the GT segmentation maps of the corresponding source images. This network is used for training ProCST. Let $Label(s_N)$ be the GT segmentation map of scale $N$, then the SiT label loss is
\begin{equation}
    \mathcal{L}_{CE,SiT}=\mathcal{L}_{CE}[Segmentor_{SiT}(\Tilde{t_N}), Label(s_N)].
\end{equation}
This loss updates both the weights of $Segmentor_{SiT}$ and ProCST. Specifically, it flows to the $G_N^{st}$ generator and helps it distinguish both the class of the current pixel and the spatial distribution of each class in the source domain. These properties help the translation process and also tend to improve segmentation properties of the translated SiT image. 

\textit{Cyclic label loss.} In addition to the above network, we present our novel cyclic label loss. Feeding both generators with segmentation loss will transfer the segmentation loss between domains. This property is very important to the UDA task, as the goal is good segmentation in the target domain. We denote the output of the cycle of source$\rightarrow$target$\rightarrow$source by $\Tilde{\Tilde{s}}$, which is a source domain image. As annotations from the source domain are available, we pretrained another ResNet101 on the source dataset, and used this pretrained $Segmentor_S$ as part of a loss for the cyclic $\mathbb{S}\xrightarrow{}\mathbb{T}\xrightarrow{}\mathbb{S}$ loop. Specifically, our loss requires the segmentation map that is trained on the real source data to succeed also on $\Tilde{\Tilde{s}}$, i.e., the loss is given by
\begin{equation}
    \mathcal{L}_{CE,ss}=\mathcal{L}_{CE}[Segmentor_{S}(\Tilde{\Tilde{s}}_N), Label(s_N)].
\end{equation}
This loss impacts both generators and acts as a segmentation information transporter between source and target domains. 
To conclude, the label loss is given by
\begin{equation}
        \mathcal{L}_{Labels} =  \lambda_{labels}(\mathcal{L}_{CE,sit} + \mathcal{L}_{CE,ss}),
\end{equation}
and thus the complete loss of scale $1 \le n \le N$ given by 
\begin{equation}
    \begin{aligned}
        &\mathcal{L}_{1 \le n < N} = \mathcal{L}_{CST}, \\
        &\mathcal{L}_{n=N} = \mathcal{L}_{CST} + \mathcal{L}_{Labels}.
    \end{aligned}   
\end{equation}

\begin{figure*}[t]
    \centering
    \begin{subfigure}{0.32\linewidth}
        \centering
        \includegraphics[width=\linewidth]{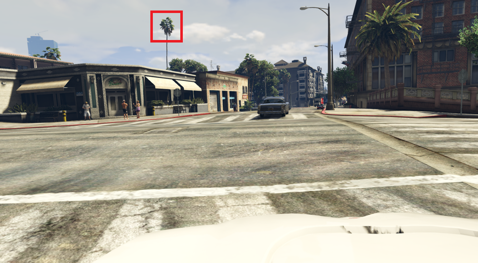}
    \end{subfigure}
    \begin{subfigure}{0.32\linewidth}
        \centering
        \includegraphics[width=\linewidth]{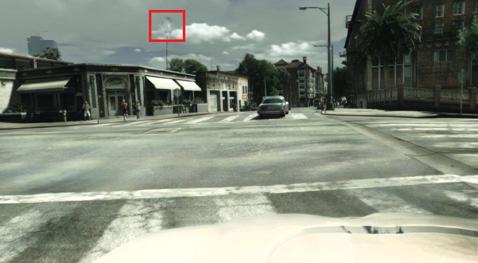}
    \end{subfigure}
    \begin{subfigure}{0.32\linewidth}
        \centering
        \includegraphics[width=\linewidth]{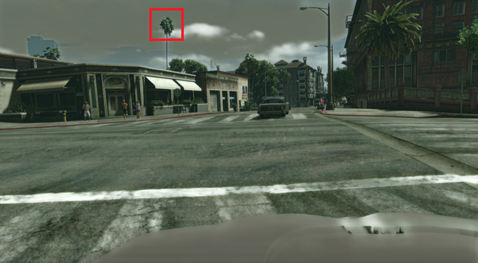}
    \end{subfigure}
    \caption{{\bf ProCST preservers the source image content.} Left: original source image. Middle: translated image from GTA5 to Cityscapes using EPE. Right: translated image from GTA5 to Cityscapes using ProCST. Note that the tree in the red box disappears after EPE's translation, but the translation of ProCST preservers the tree although it is surrounded entirely by sky.}
    \label{fig:epe_compare}
\end{figure*}

\section{Experiments}
\label{sec:experiment}
We turn to present the results of our proposed framework. We first show the results of integrating three robust UDA methods within our framework. We also provide a qualitative comparison between the images in the SiT data and the original source data and a quantitative comparison between the results achieved using ProCST SiT data and those achieved using data generated by other state-of-the-art image-to-image translation methods. Finally, we present an ablation study that demonstrates the importance of the different parts in the ProCST model. In the appendix, we present the implementation details of the ProCST model, qualitative comparison between ProCST's translation and other image-to-image translation methods, and additional outputs of ProCST.

\paragraph{Results}
We trained our model on two source datasets, GTA5 and Synthia, that were then translated to obtain the new SiT data (created per dataset). The data is used to train three state-of-the-art UDA methods, ProDA \cite{zhang2021prototypical}, DAFormer \cite{hoyer2021daformer} and HRDA \cite{hoyer2022hrda}. 

Table~\ref{table:results_gta_synthia} shows the results compared to other methods and also to HRDA, DAFormer and ProDA when trained with the regular source data.
Notice that training ProDA using our SiT data resulted in an improvement of 1.1\% mIoU over the original GTA5 dataset, and improvement of 0.6\% mIoU over the original Synthia dataset. Training DAFormer using our SiT dataset achieved improvement of 1.1\% mIoU over the original GTA5 dataset, and improvement of 0.9\% mIoU over the original Synthia dataset. Training HRDA using our SiT dataset achieved improvement of 1.1\% mIoU over the original GTA5 dataset, and improvement of 0.8\% mIoU over the original Synthia dataset. Note that the results of DAFormer and HRDA refer to our runs, on 8 seeds. The SiT data generated by ProCST improves the accuracy of three state-of-the-art UDA methods on both GTA5 and Synthia datasets.

Figure \ref{fig:sit_tis_translations} shows the similarity between images from the SiT data, generated by ProCST, to images from the target data. The GTA5 data tries to simulate the view in the yellow and sunny Los Angeles, while the Cityscapes data was shot in the roads of the green and cloudy Germany. We can see how the ProCST model catches these differences and translates the yellow-red images of GTA5 onto images with blue-green atmosphere, while the translation in the opposite direction gives the image a red-yellow atmosphere.

In addition, we can see that ProCST changes appearance and textures.
The colors and textures of the road turn from rough yellow-grey in GTA5 to smooth gray, as in Cityscapes. The sky turns from clear blue to foggy and grey. The color of the vegetation turns from green-yellow into deep green. These changes appear also in the opposite translation. While the style and appearance of the images change, the visual content and details do not. This is an important property for UDA tasks, as annotations are available only on the original source dataset. Furthermore, we can see similar trends also in the Synthia to Cityscapes translation.

\begin{table}
\small
    \caption{{\bf Comparison to other SiT generation options.} We used existing UDA and image-to-image translation methods to generate SiT data for the training of DAFormer.
    Double horizontal line indicates that results were tested on the exact subset of GTA5 dataset that was publicly available. Units of mIoU are \%. We use the same seed in all experiments for a fair comparison. Notice how ProCST leads to a better improvement of DAFormer in all cases.} 
    \centering
    \label{table:sit_compare_table}
     {\renewcommand{\arraystretch}{1.2}
    \setlength{\tabcolsep}{.1em}
    \begin{tabular}{|| c |  c || c | c  || c | c || c | c || }
    \hline 
    \text{SiT} & \text{mIoU} & \text{SiT} & \text{mIoU} & \text{SiT} & \text{mIoU} & \text{SiT} & \text{mIoU}\\ 
    \hline
    \makecell{FDA}      & 67.2 & \makecell{CUT}      & 64.3 & \makecell{CyGAN}    & 65.9 &  
    \makecell{EPE}    & 63.3\\
    \makecell{BDL}      & 67.0 & \makecell{MUINT}      & 64.4 & \makecell{Source}      & 67.4 &
    \makecell{Source}      & 66.5 \\
    \makecell{Source}      & 67.0 & \makecell{Source}      & 66.9 & \makecell{\bf{ProCST}}   &  \textbf{68.8} &
    \makecell{\bf{ProCST}}   &  \textbf{69.1}\\
    \makecell{\bf{ProCST}}   &  \textbf{68.7} & \makecell{\bf{ProCST}}   &  \textbf{68.5} & & & & \\
    \hline 
    \end{tabular}
    }
\end{table}

\begin{table*}[t]
\small
\centering
    \caption{Comparison to state-of-the-art UDA methods for semantic segmentation. Double vertical line separates the results of $\text{GTA5} \rightarrow \text{Cityscapes}$ (above the line), and the results of $\text{Synthia} \rightarrow \text{Cityscapes}$ (below the line). *Average of 8 seeds, results using our runs.} \label{table:results_gta_synthia}
    {\renewcommand{\arraystretch}{1.05}
    \setlength{\tabcolsep}{.12em}
    \begin{tabular}{ c | c  c  c  c  c  c  c  c  c  c   c  c  c  c  c  c  c  c  c | c }
        & \rotatebox[origin=c]{90}{Road} & \rotatebox[origin=c]{90}{S.walk} & \rotatebox[origin=c]{90}{Build.} & \rotatebox[origin=c]{90}{Wall} & \rotatebox[origin=c]{90}{Fence} & \rotatebox[origin=c]{90}{Pole} & \rotatebox[origin=c]{90}{Light} & \rotatebox[origin=c]{90}{Sign} & \rotatebox[origin=c]{90}{Vege.} & \rotatebox[origin=c]{90}{Terrain} & \rotatebox[origin=c]{90}{Sky} & \rotatebox[origin=c]{90}{Person} & \rotatebox[origin=c]{90}{Rider} & \rotatebox[origin=c]{90}{Car} & \rotatebox[origin=c]{90}{Truck} & \rotatebox[origin=c]{90}{Bus} & \rotatebox[origin=c]{90}{Train} & \rotatebox[origin=c]{90}{M.bike} & \rotatebox[origin=c]{90}{Bike} & \rotatebox[origin=c]{90}{\textbf{mIoU}} \\ \hline 
         
         
        \text{CyCADA} \shortcite{hoffman2018cycada}                          & 86.7 & 35.6 & 80.1 & 19.8 & 17.5 & 38.0   & 39.9 & 41.5 & 82.7 & 27.9 & 73.6 & 64.9 & 19.0   & 65.0 & 12.0 & 28.6 & 4.5  & 31.1 & 42.0  & 42.7\\  
        \text{CBST}	\shortcite{park2019semantic}                            & 91.8 & 53.5 & 80.5 & 32.7 & 21.0 & 34.0 & 28.9 & 20.4 & 83.9 & 34.2 & 80.9 & 53.1 & 24.0 & 82.7 & 30.3 & 35.9 & 16.0 & 25.9 & 42.8 & 45.9 \\
        \text{FADA} \shortcite{wang2020classes}                            & 91.0 & 50.6 & 86.0 & 43.4 & 29.8 & 36.8 & 43.4 & 25.0 & 86.8 & 38.3 & 87.4 & 64.0 & 38.0 & 85.2 & 31.6 & 46.1 & 6.5  & 25.4 & 37.1 & 50.1 \\						
        \text{DACS}	\shortcite{tranheden2021dacs}                            & 89.9 & 39.7 & 87.9 & 30.7	& 39.5 & 38.5 & 46.4 & 52.8	& 88.0 & 44.0 & 88.8 & 67.2 & 35.8 & 84.5 & 45.7 & 50.2 & 0.0 & 27.3 & 34.0 & 52.1 \\
        \text{CorDA} \shortcite{wang2021domain}                            & 94.7 & 63.1 & 87.6 & 30.7 & 40.6 & 40.2 & 47.8 & 51.6 & 87.6 & 47.0 & 89.7 & 66.7 & 35.9 & 90.2 & 48.9 & 57.5 & 0.0 & 39.8 & 56.0 & 56.6 \\ 
        \hline
        \text{ProDA} \shortcite{zhang2021prototypical}                    & 87.8 & 56.0 & 79.7 & \textbf{46.3} & 44.8 & 45.6 & 53.5 & 53.5 & 88.6 & 45.2 & \textbf{82.1} & 70.7 & \textbf{39.2} & 88.8 & 45.5 & \textbf{59.4} & \textbf{1.0} & 48.9 & 56.4 & 57.5 \\
        \makecell{ProDA+ProCST}       & \textbf{90.0} & \textbf{61.2} & \textbf{81.2} & 40.8 & \textbf{45.9} & \textbf{49.0} & \textbf{57.8} & \textbf{59.8} & \textbf{88.9} & \textbf{46.9} & 80.1 & \textbf{72.8} & 34.6 & \textbf{89.5} & \textbf{46.3} & 58.7 & 0.0  & \textbf{51.0} & \textbf{59.2} & \textbf{58.6} \\ \hline
        
        \text{DAFormer*} \shortcite{hoyer2021daformer} & 95.1 & 67.6 & 89.3 & 53.3 & 44.9 & 48.8 & 56.0 & 60.4 & 89.8 & 47.5 & 92.0 & 71.8 & 44.8 & 92.0 & 70.1 & 78.4 & 64.4 & 55.7 & 62.8 & 67.6  \\
        \makecell{DAFormer*+ProCST}  & \textbf{95.8} & \textbf{69.6} & \textbf{89.8} & \textbf{55.8} & \textbf{45.0} & \textbf{49.8} & \textbf{56.8} & \textbf{63.3} & \textbf{90.2} & \textbf{50.3} & \textbf{93.0} & \textbf{72.2} & \textbf{44.9} & \textbf{92.3} & \textbf{72.2} & \textbf{78.8} & \textbf{65.1} & \textbf{56.4} & \textbf{63.1} & \textbf{68.7}  \\ \hline
        
        \text{HRDA*} \shortcite{hoyer2022hrda} & 96.0 & 72.8 & 91.2 & \textbf{60.8} & 51.0 & 58.2 & 63.6 & 71.5 & 91.5 & 50.3 & \textbf{93.9} & 78.4 & 51.1 & 93.8 & 80.1 & 84.4 & 63.8 & 63.7 & 67.6 & 72.8    \\
        \makecell{HRDA*+ProCST}  &  \textbf{96.7} & \textbf{75.8} & \textbf{91.3} & 60.3 & \textbf{52.8} & \textbf{58.8} & \textbf{66.0} & \textbf{72.1} & 91.5 & \textbf{50.4} & 93.8 & \textbf{78.6} & \textbf{52.1} & \textbf{94.0} & \textbf{81.4} & \textbf{85.7} & \textbf{71.8} & \textbf{63.8} & 67.6 & \textbf{73.9} \\

        \hline \hline
        \text{CBST}    \shortcite{park2019semantic}                         & 68.0 & 29.9 & 76.3 & 10.8 & 1.4 & 33.9 & 22.8 & 29.5 & 77.6 & - & 78.3 & 60.6 & 28.3 & 81.6 & - & 23.5 & - & 18.8 & 39.8 & 42.6 \\
        \text{DACS}   	\shortcite{tranheden2021dacs}                           & 80.6 & 25.1 & 81.9 & 21.5 & 2.9 & 37.2 & 22.7 & 24.0 & 83.7 & - & 90.8 & 67.6 & 38.3 & 82.9 & - & 38.9 & - & 28.5 & 47.6 & 48.3 \\
        \text{CorDA}  \shortcite{wang2021domain}                           & 93.3 & 61.6 & 85.3 & 19.6 & 5.1 & 37.8 & 36.6 & 42.8 & 84.9 & - & 90.4 & 69.7 & 41.8 & 85.6 & - & 38.4 & - & 32.6 & 53.9 & 55.0 \\ 
        
        \hline
        \text{ProDA}    \shortcite{zhang2021prototypical}                        & 87.8 & 45.7 & 84.6 & \textbf{37.1} & 0.6 & 44   & 54.6 & 37.0 & \textbf{88.1} & - & 84.4 & \textbf{74.2} & 24.3 & 88.2 & - & 51.1 & - & \textbf{40.5} & \textbf{45.6} & 55.5 \\
        \makecell{ProDA+ProCST}       & 87.8 & \textbf{48.2} & \textbf{85.8} & 22.9 & \textbf{0.7}  & \textbf{47.1} & \textbf{56.4} & \textbf{47.1} & 88.0 & - & \textbf{86.8} & 72.4 & \textbf{25.4} & \textbf{90.2} & - & \textbf{58.0} & - & 38.3 & 41.9 & \textbf{56.1} \\ \hline

        \text{DAFormer*} \shortcite{hoyer2021daformer} & \textbf{84.4} & \textbf{41.8} & \textbf{87.9} & 40.4 & \textbf{6.9} & 48.8 & 54.0 & 53.7 & 86.0 & - & \textbf{89.0} & 72.5 & 45.6 & 86.6 & - & 58.3 & - & 53.1 & 59.6 & 60.5  \\
        \makecell{DAFormer*+ProCST}  & 84.3 & 41.1 & 87.7 & \textbf{42.6} & 6.1 & \textbf{50.7} & \textbf{55.5} & \textbf{54.2} & \textbf{86.1} & - & 87.9 & \textbf{74.7} & \textbf{47.2} & \textbf{87.6} & - & \textbf{61.4} & - & \textbf{53.3} & \textbf{62.5} & \textbf{61.4} \\ \hline
        
        \text{HRDA*} \shortcite{hoyer2022hrda} & 84.5 & 43.9 & 88.7 & 50.0 & \textbf{7.0} & 57.3 & 64.8 & 60.5 & 86.1 & - & 92.9 & \textbf{79.1} & 52.3 & \textbf{89.6} & - & \textbf{65.3} & - & 63.3 & 64.3 & 65.6   \\
        \makecell{HRDA*+ProCST}  & \textbf{86.7} & \textbf{50.7} & 88.7 & 50.0 & 6.4 & \textbf{58.0} & \textbf{66.6} & \textbf{62.2} & \textbf{86.4} & - & \textbf{93.6} & 78.8 & \textbf{52.7} & 88.4 & - & 64.6 & - & \textbf{63.5} & \textbf{64.4} & \textbf{66.4} \\ \hline
    \end{tabular}
    }
\end{table*}

\paragraph{Comparison of ProCST to other image-to-image translation methods}
Other state of the art image-to-image translation methods lack content preservation property and thus do not achieve as good results on UDA tasks. For instance, we can see the difference between image translation of the state-of-the-art translation approach Enhancing Photorealism Enhancement, EPE \cite{richter2021enhancing}, and the image translation of ProCST in Figure~\ref{fig:epe_compare}. While EPE's method has very good visual quality, it sometimes deletes some of the content of the original image, such as trees surrounded by sky background. The combination of cyclic loss together with style loss helps ProCST to change the style but yet preserve the content, which is a crucial property to the success in a generic UDA task. More examples can be found in the appendix. Although EPE's translation performs well on image visual quality, it does not achieve good accuracy when used as a SiT data generator for UDA, as shown in Table~\ref{table:sit_compare_table}. 

Additionally, we have compared ProCST to other image-to-image translation methods when used as SiT data generators: BDL \cite{li2019bidirectional}, CUT \cite{park2020contrastive}, CycleGAN \cite{zhu2017unpaired}, MUINT \cite{huang2018munit} and FDA \cite{yang2020fda}. We chose a large variety of translation methods for robust comparison. Qualitative comparison of the translations outputs appears in the appendix. As Table~\ref{table:sit_compare_table} clearly shows, a mere preprocessing of the data with other methods does not improve significantly UDA accuracy in the semantic segmentation task. The architecture and losses of ProCST (such as the cyclic label loss, which is novel in UDA context) that we proposed are key for allowing the improvement achieved.

\paragraph{Ablation study}
We trained a ProCST model using 5 configurations to find out what parts of the proposed architecture contribute most to the boost in performance. The examined parameters are the multiscale pyramid length (controlled by $N$), label loss term (controlled by $\lambda_{labels}$), style loss term (controlled by $\lambda_{style}$) and cyclic loss term (controlled by $\lambda_{cyclic}$). Each ablation discards one parameter, namely, set either number of levels $N$ to 1 or the corresponding $\lambda$ to 0.
We perform the ablation using the DAFormer on GTA5 $\rightarrow$ Cityscapes with the SiT data as input and a minor hyper parameter tuning by changing DAFormer's EMA update parameter to $\alpha=0.997$ from the original value of $\alpha=0.999$. The reason for this fine tuning lays on the differences between the source and the SiT datasets. 
For a complete and fair comparison, we also show the results of DAFormer using the original GTA5 dataset as source with both values of $\alpha$. 
Table~\ref{table:ablation_on_daformer} shows that if we omit either the style or cyclic loss terms, ProCST fails to converge. Moreover, omitting either the multiscale structure or label loss results in performance degradation compared to the full ProCST model. Note that we use just one (same) seed in the ablation (thus, numbers slightly differ from Table~\ref{table:results_gta_synthia}).
\begin{table}
\centering
    \caption{DAFormer trained on different ProCST generated SiT datasets. 
    Training configuration of DAFormer is as its original configuration, excluding the EMA update parameter $\alpha$. We change it $\alpha=0.997$ in rows marked with *.} 
    \label{table:ablation_on_daformer}
    {\renewcommand{\arraystretch}{1.2}
    \setlength{\tabcolsep}{.2em}
    \begin{tabular}{ c | c c c c | c c }
    \text{Dataset} & \makecell{Label \\ Loss} & \makecell{Multiscale \\ Pyramid} & \makecell{Style \\ Loss} & \makecell{Cyclic \\ Loss} & \text{mIoU[\%]} & \text{Rel.} \\ 
    \hline
    Source \     &   &   &  &  & 66.4 & +0.0 \\
    Source*      &   &   &  &  & 66.3 & -0.1 \\ 
    \hline
                                & \checkmark & \checkmark & $\times$    &  \checkmark   & $\times$ & $\times$ \\
                                & \checkmark & \checkmark & \checkmark  & $\times$      & $\times$ & $\times$ \\
    \makecell{ProCST*}          & $\times$   & \checkmark & \checkmark  &  \checkmark   & 67.2 & +0.8 \\
                                & \checkmark & $\times$ &  \checkmark &    \checkmark   & 67.2 & +0.8 \\
                                & \checkmark & \checkmark & \checkmark  &  \checkmark   & \textbf{68.2} & \textbf{+1.8} \\
     \hline
    \end{tabular}
    }
\end{table}

\section{Conclusion}
In this work, we have shown a new concept that achieves improved performance of generic UDA methods. Our image-to-image translation method narrows the domain gap as a prior step to UDA training. The unique loss combination is tailored to preserve image content yet change its style. We have shown that this improves the ProDA, DAFormer and HRDA state-of-the-art UDA methods. To our knowledge, our cyclic label loss was not used before. From the examination we performed in this work, one can observe that this term regularizes the training and allows the model to achieve superior translation quality. We believe that this concept can be extended to other types of problems, where annotating images in a target domain of interest is very costly.

\noindent {\bf Acknowledgment.} This research was supported by ERC-StG SPADE grant no. 757497. We would like to thank Deborah Cohen for her helpful comments.

\appendix
\section{Appendix}
In the following sections, we provide additional qualitative results and explanations that elaborate more deeply the unique contribution of ProCST as a booster to generic UDA tasks. 
First, we provide implementation details regarding ProCST architecture and training process. Then, we present a detailed comparison between ProCST SiT data and EPE \cite{richter2021enhancing} generated data. Additionally, we provide wide qualitative comparison between the translation of ProCST and this of other image-to-image translation methods. Those comparisons illustrate the content preservation property of ProCST, which is crucial for narrowing the domain gap. Moreover, we discuss the qualitative differences between images generated using all the models presented in the ablation study presented in Section~\ref{sec:experiment}. Finally, we present more examples for SiT and TiS translated images, generated using both GTA5 and Synthia as the source datasets.

\subsection{Implementation Details}

\textbf{Datasets.} Source datasets: The GTA5 dataset \cite{GTA5} contains 24,966 images of size (1052, 1914) and their pixel-wise annotations. The Synthia dataset contains 9400 images of size (760, 1280) and their pixel-wise annotations.

Target dataset: The Cityscapes dataset \cite{cityscapes}, with the extra available train images: total of 22,973 images of size (1024, 2048). 

Prepossessing of the source datasets includes two steps: first, we resize the input images to a size that preserves aspect ratio of the original source image. This size varies between domains (different domains have different image size and aspect ratio), but all domain resized images are larger than (512, 1024). Then, we take a random crop of size (512, 1024) from the resized images. The target dataset has an aspect ratio of 1:2. Therefore, its prepossessing includes only resizing to (512, 1024), without any cropping. Thus, $(H_N, W_N)=(512, 1024)$.

\textbf{Network Architecture.}
We use $N=3$ scales, with a scale factor $r=0.5$. Generators and discriminators are fully convolutional, with width of 64 channels for all $n$ and depth of 5 layers at scale $n=1$, and 7 layers at scales $n=2,3$. For all generators and discriminators, we used a normalization according to the current scale's batch size. Following \citet{wu2018group}, we normalized using Batch Normalization only  if the number of images per GPU in the current scale exceeds 16. Otherwise, we used Group Normalization with groups number $G=8$.
We used DeepLabV2 \cite{chen2017deeplab} for both segmentation networks.

\textbf{Training.} 
We used the Adam optimizer for all generators and discriminators networks with parameters $lr_g,lr_d=0.0001$. The optimizer for the segmentation networks is SGD with learning rate of $lr_{semseg}=0.0001$.
The weights of the losses were chosen to prefer style transfer, but if we would have discarded one of the other losses it would have caused either a dramatic performance degradation or a lack of model convergence, as can be seen in Section~\ref{sec:experiment}.
We adopt the values of $\lambda_{content}, \lambda_{style,Gram}$ and $\lambda_{TV}$ from \citet{johnson2016perceptual}. To control the relative size of this loss compared to the other losses in ProCST, we normalized the maximum weight to 1 and kept the ratio between all other weights. The original values are $\lambda_{content}=1, \lambda_{style,Gram}=30, \lambda_{TV}=1$.
For $\mathcal{L}_{CST}$ we set $\lambda_{adv.}=1, \lambda_{style}=10, \lambda_{cyclic}=1$, and for the label loss we set $\lambda_{labels}=3$.

\textbf{SiT Data Creation.} After training ProCST as described above, we generate the SiT data, where the input of the ProCST model are images from GTA5 or Synthia, and the output image are in Cityscapes style, as figures \ref{fig:sit_tis_gta_translations} and \ref{fig:sit_tis_synthia_translations} show. Note that we use full resolution source images without any preprocessing, thus ProCST's output has the exact shape of the original source images and annotations. ProCST is fully convolutional and thus can generate SiT data with varying input shape. We train ProCST with maximum resolution of 1024x512, and use this pretrained ProCST model to generate images with varying resolution according to the full resolution of the source images. 

\subsection{Comparison of ProCST and EPE}
\begin{figure*}
    \centering
    \begin{subfigure}{0.32\linewidth}
        \centering
        \includegraphics[width=\linewidth]{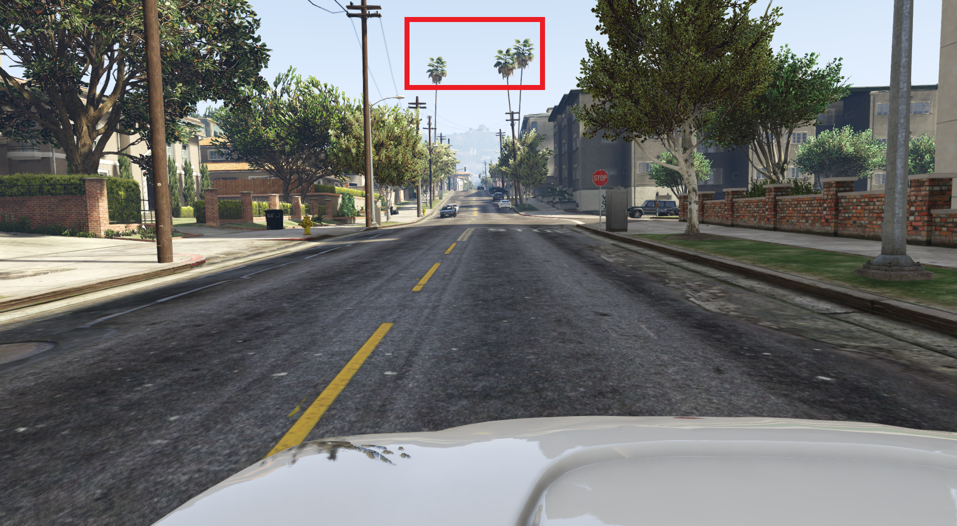}
    \end{subfigure}
    \begin{subfigure}{0.32\linewidth}
        \centering
        \includegraphics[width=\linewidth]{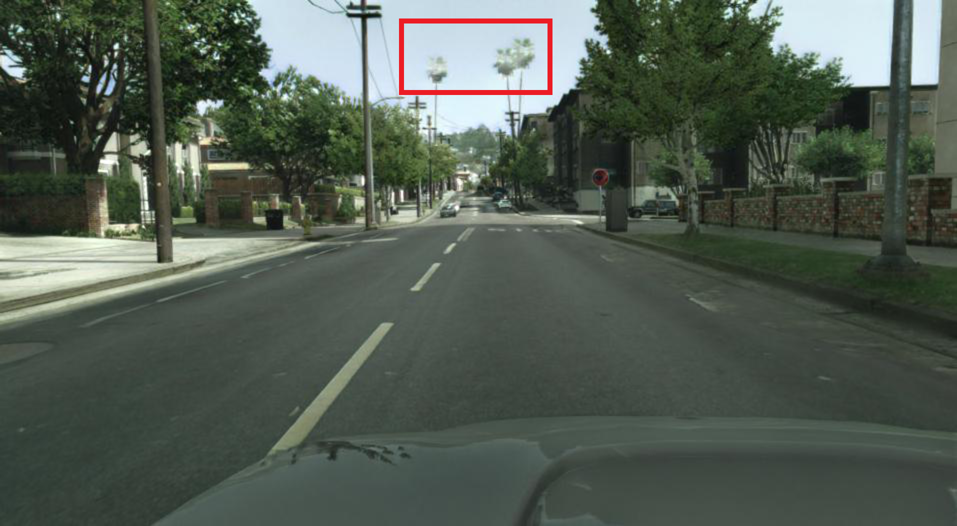}
    \end{subfigure}
    \begin{subfigure}{0.32\linewidth}
        \centering
        \includegraphics[width=\linewidth]{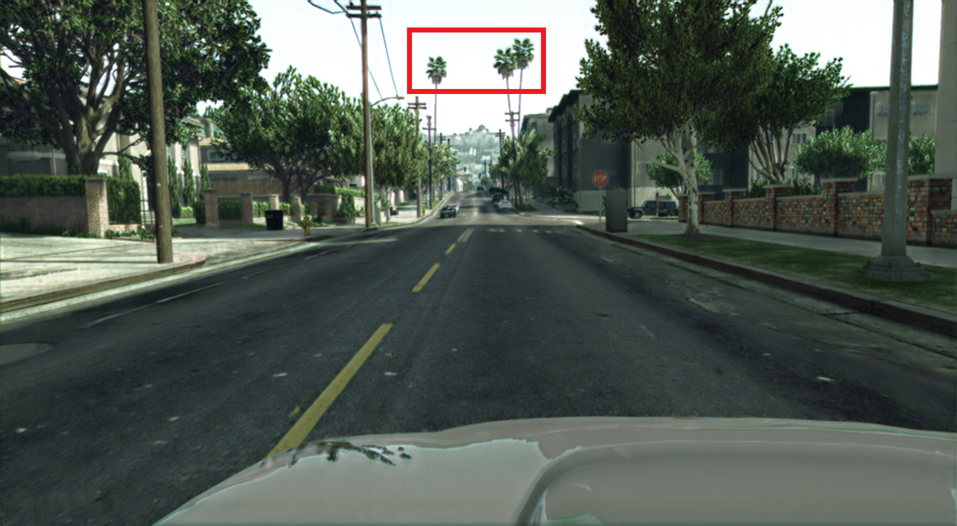}
    \end{subfigure}
    
    \begin{subfigure}{0.32\linewidth}
        \centering
        \includegraphics[width=\linewidth]{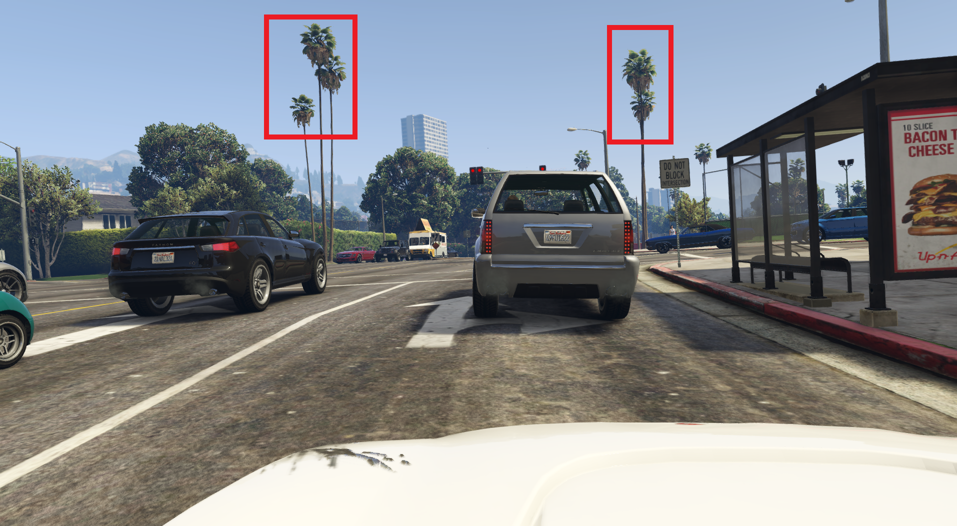}
    \end{subfigure}
    \begin{subfigure}{0.32\linewidth}
        \centering
        \includegraphics[width=\linewidth]{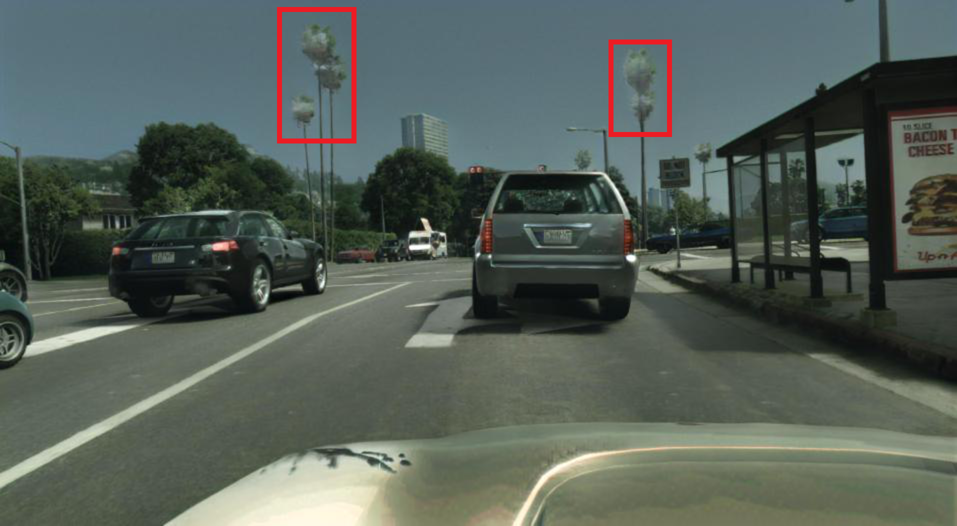}
    \end{subfigure}
    \begin{subfigure}{0.32\linewidth}
        \centering
        \includegraphics[width=\linewidth]{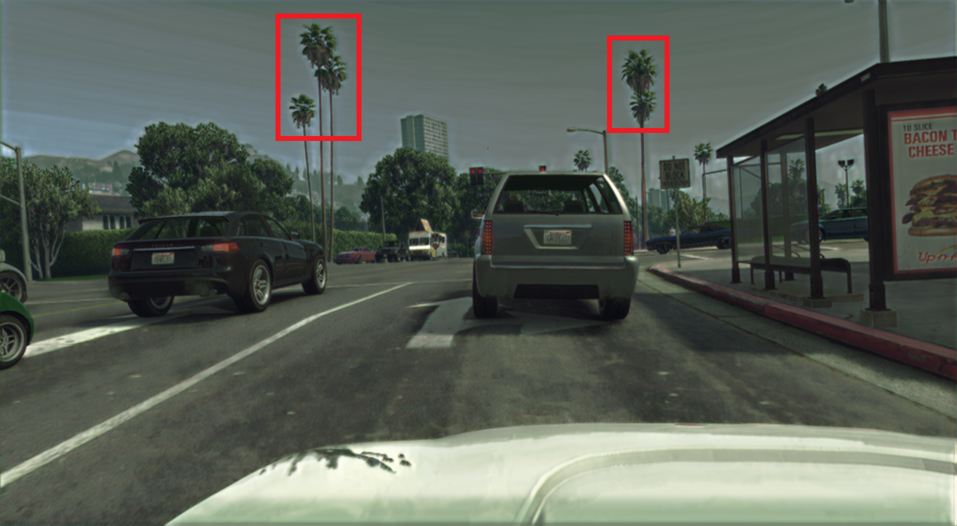}
    \end{subfigure}
    \begin{subfigure}{0.32\linewidth}
        \centering
        \includegraphics[width=\linewidth]{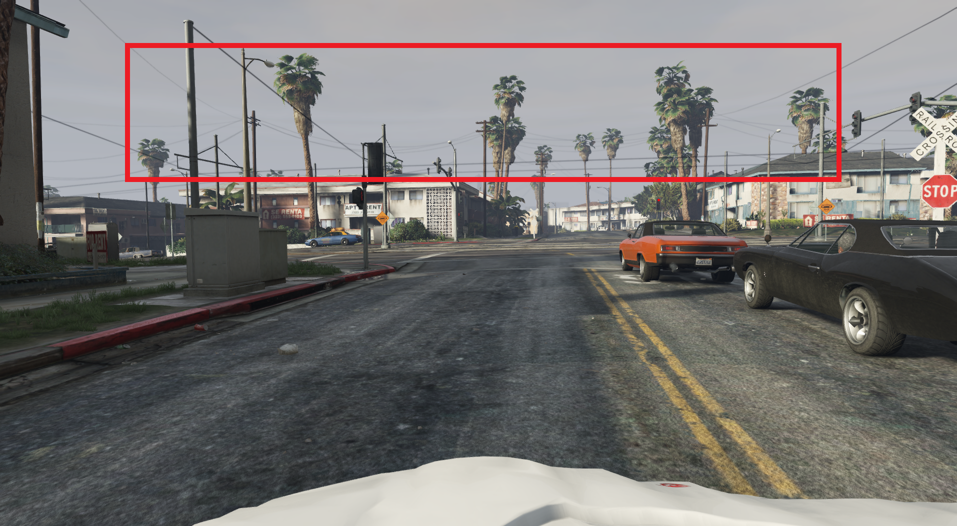}
    \end{subfigure}
    \begin{subfigure}{0.32\linewidth}
        \centering
        \includegraphics[width=\linewidth]{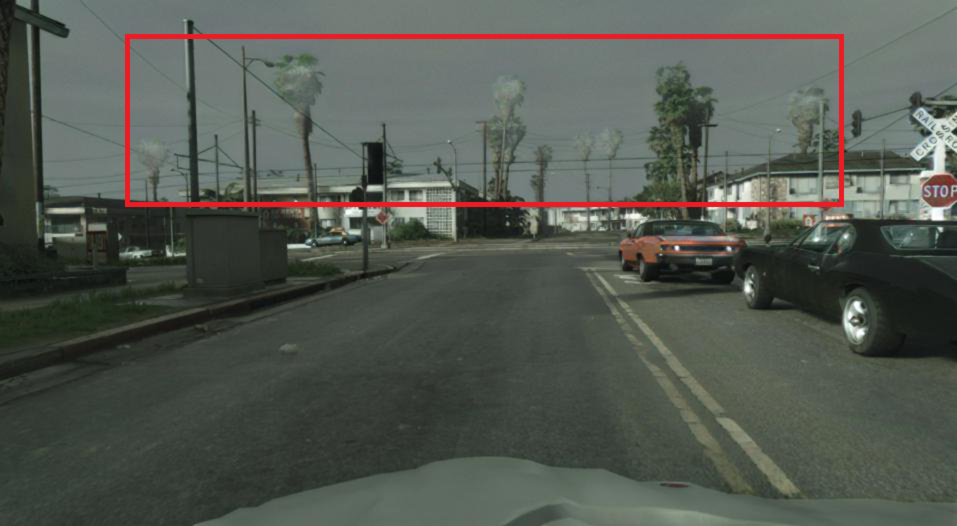}
    \end{subfigure}
    \begin{subfigure}{0.32\linewidth}
        \centering
        \includegraphics[width=\linewidth]{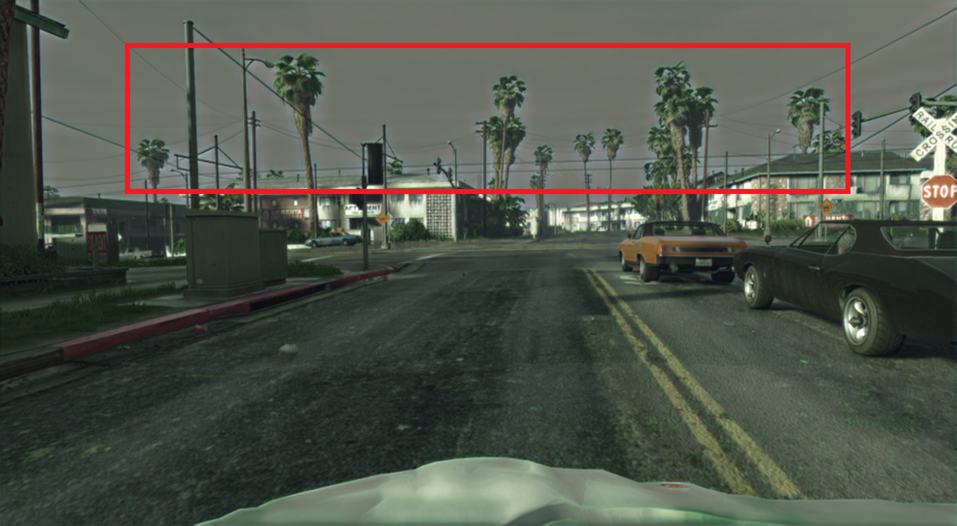}
    \end{subfigure}
    \caption{\textbf{Content Preservation.} Left to right: Original source image; EPE translated image; ProCST translated image. ProCST's output results in a clean image while EPE \cite{richter2021enhancing} distorts the palm trees due to the surrounding sky.}
    \label{fig:epe_compare_sky}
\end{figure*}

We turn to present more comparisons between ProCST and EPE. Figure~\ref{fig:epe_compare_sky} presents more examples for objects that EPE's translation deletes or distorts, while ProCST's translation preserves. Figure~\ref{fig:epe_compare_sky} shows that trees in the sky are sometimes distorted due to the EPE translation. Note that these examples imply a more general trend. The EPE image translation is focused on generating good visual image quality, but not on preserving the content of the image. Thus, it sometimes distorts objects that do not visually fit their close spatial space. For example, the long palm trees are not consistent with their spatial environment (i.e. sky) and thus distorted. 

\begin{figure*}
    \centering
    \begin{subfigure}{0.7\linewidth}
        \centering
        \includegraphics[width=\linewidth]{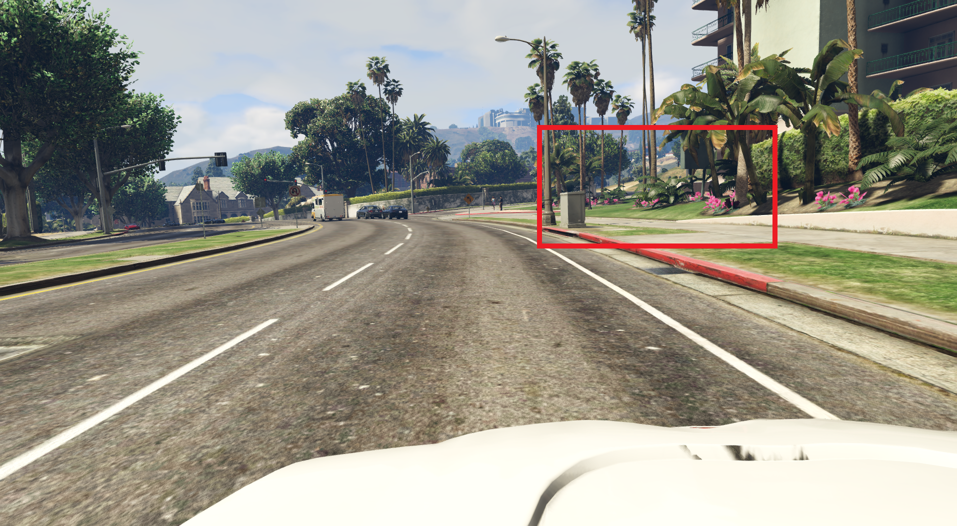}
    \end{subfigure}
    \begin{subfigure}{0.32\linewidth}
        \centering
        \includegraphics[width=\linewidth]{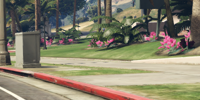}
    \end{subfigure}
    \begin{subfigure}{0.32\linewidth}
        \centering
        \includegraphics[width=\linewidth]{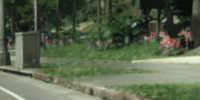}
    \end{subfigure}
    \begin{subfigure}{0.32\linewidth}
        \centering
        \includegraphics[width=\linewidth]{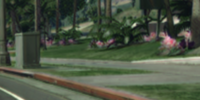}
    \end{subfigure}
    \caption{\textbf{Content Preservation.} Top row: original source image; Bottom row, left to right: Crop taken from the original image; Crop taken from EPE translated image; Crop taken from ProCST translated image. ProCST's output results in a clean image while EPE \cite{richter2021enhancing} distorts the sidewalk due to the surrounding vegetation.}
    \label{fig:epe_compare_vegetation}
\end{figure*}

We can see such a trend also in other cases. For example, in Figure~\ref{fig:epe_compare_vegetation}, we can see a crop from the original source image that contains a sidewalk surrounded by vegetation. While the EPE translation distorts the sidewalk because of the vegetation from both of its sides, the ProCST model preserves the content and results in a cleaner image that has matching content to that of the original source image.

Content preservation is a crucial property in image translation when used for domain adaptation. EPE fails to achieve this property, and thus achieves poorer performance when employed as a booster to the UDA semantic segmentation task.

\subsection{Comparison of ProCST and Additional Image-To-Image Translation Methods}
\begin{figure*}
    \begin{subfigure}{0.5\linewidth}
        \centering
        \includegraphics[width=\linewidth]{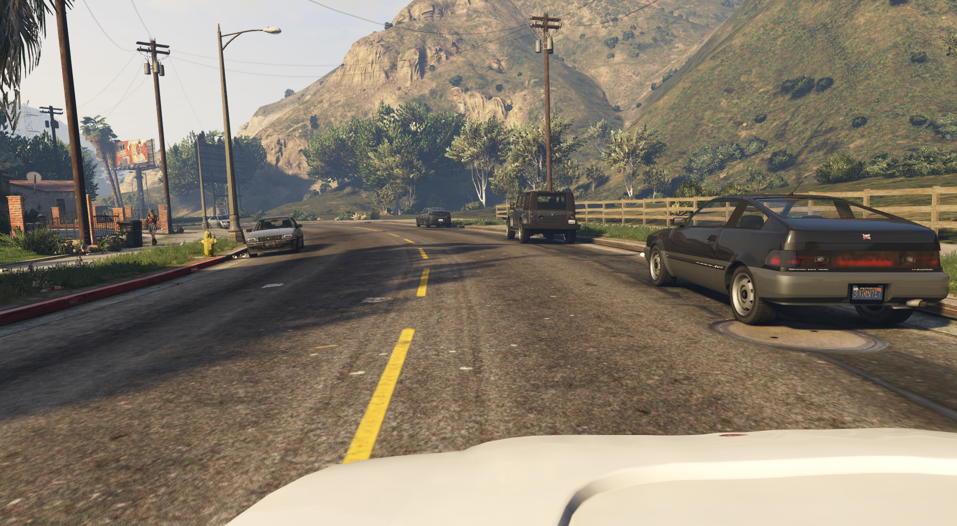}
    \end{subfigure}
    \begin{subfigure}{0.5\linewidth}
        \centering
        \includegraphics[width=\linewidth]{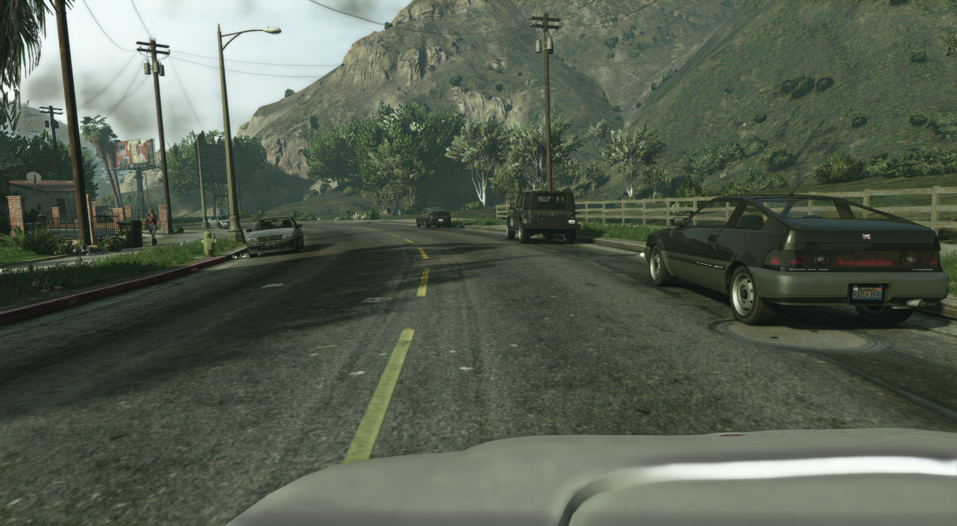}
    \end{subfigure}
        \begin{subfigure}{0.5\linewidth}
        \centering
        \includegraphics[width=\linewidth]{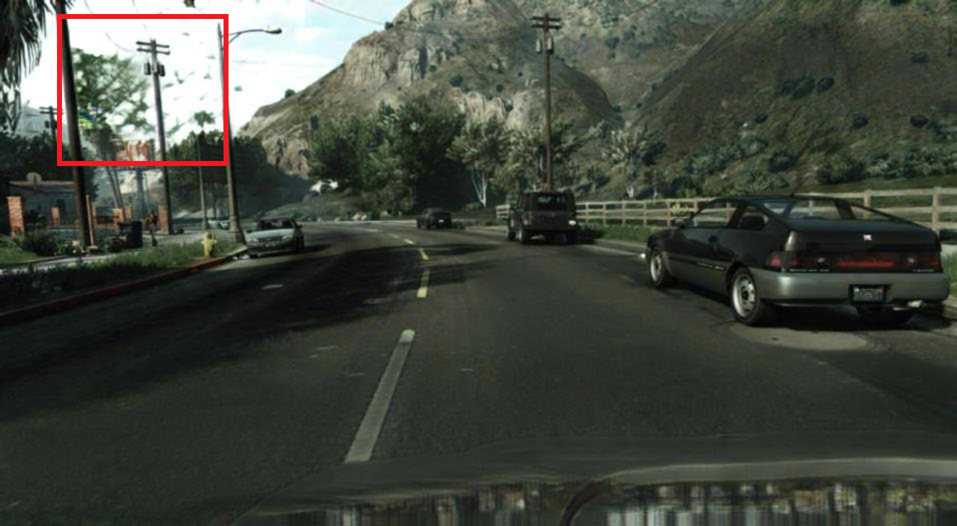}
    \end{subfigure}
    \begin{subfigure}{0.5\linewidth}
        \centering
        \includegraphics[width=\linewidth]{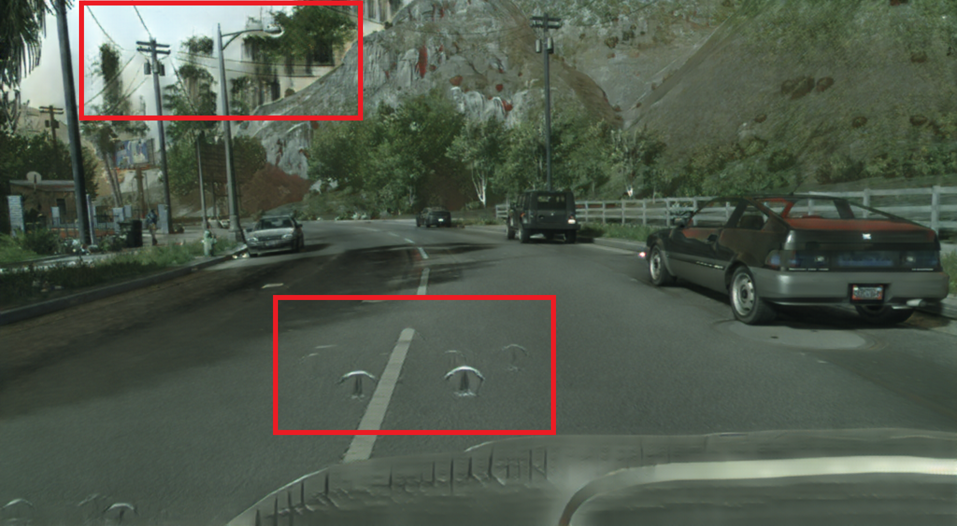}
    \end{subfigure}
    \begin{subfigure}{0.5\linewidth}
        \centering
        \includegraphics[width=\linewidth]{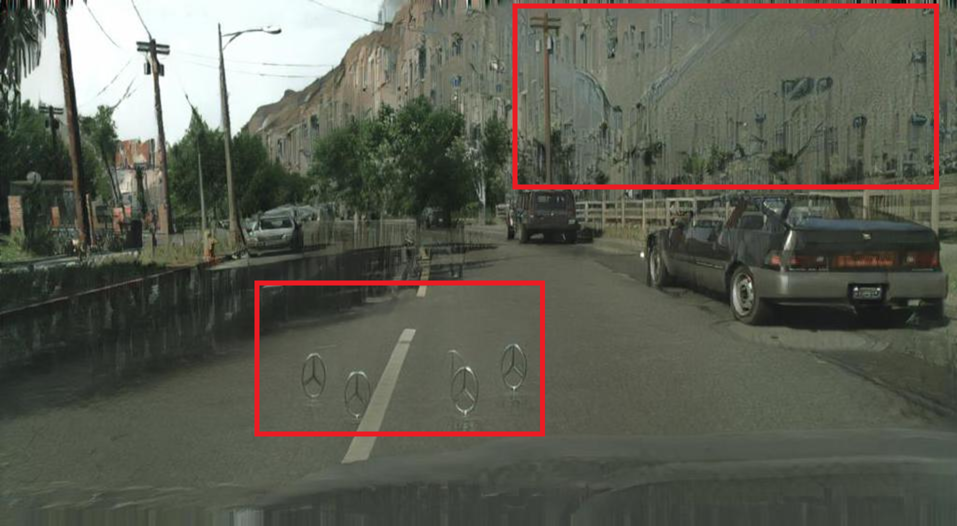}
    \end{subfigure}
    \begin{subfigure}{0.5\linewidth}
        \centering
        \includegraphics[width=\linewidth]{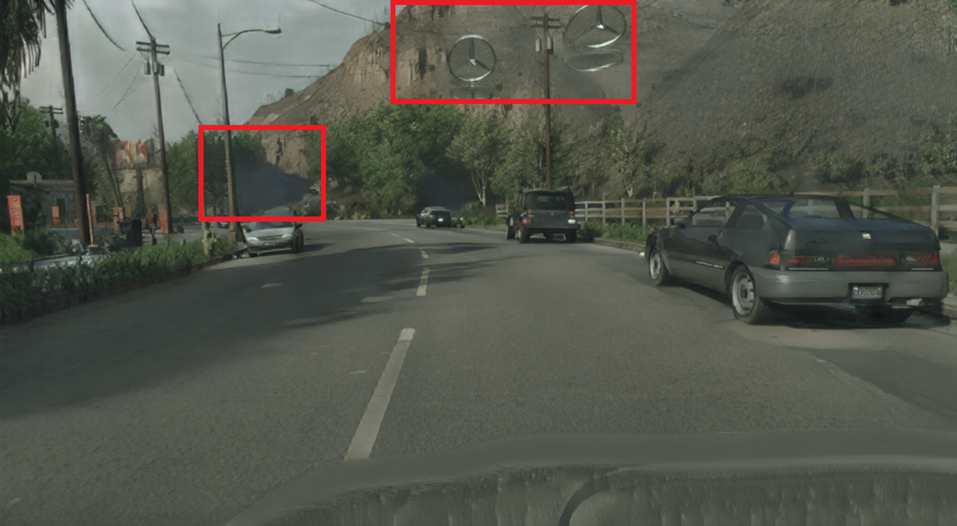}
    \end{subfigure}
    \begin{subfigure}{0.5\linewidth}
        \centering
        \includegraphics[width=\linewidth]{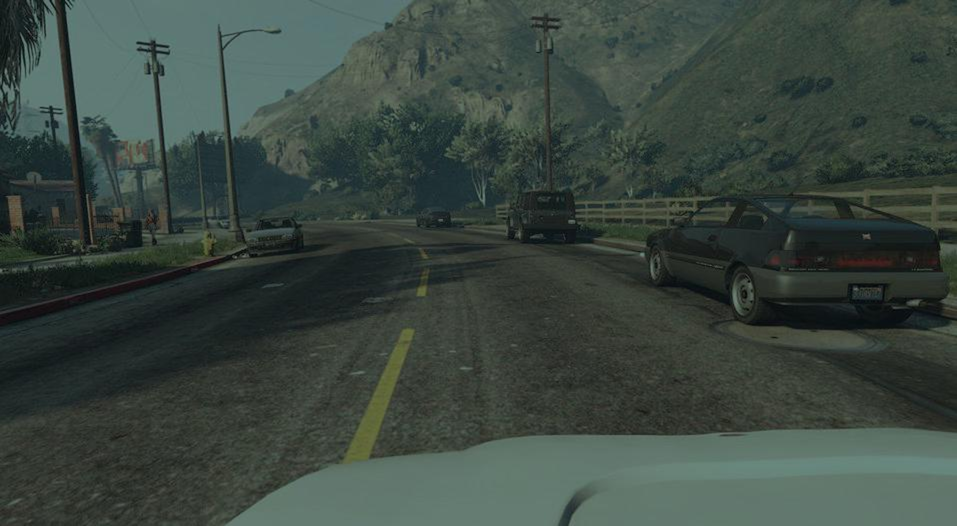}
    \end{subfigure}
    \begin{subfigure}{0.5\linewidth}
        \centering
        \includegraphics[width=\linewidth]{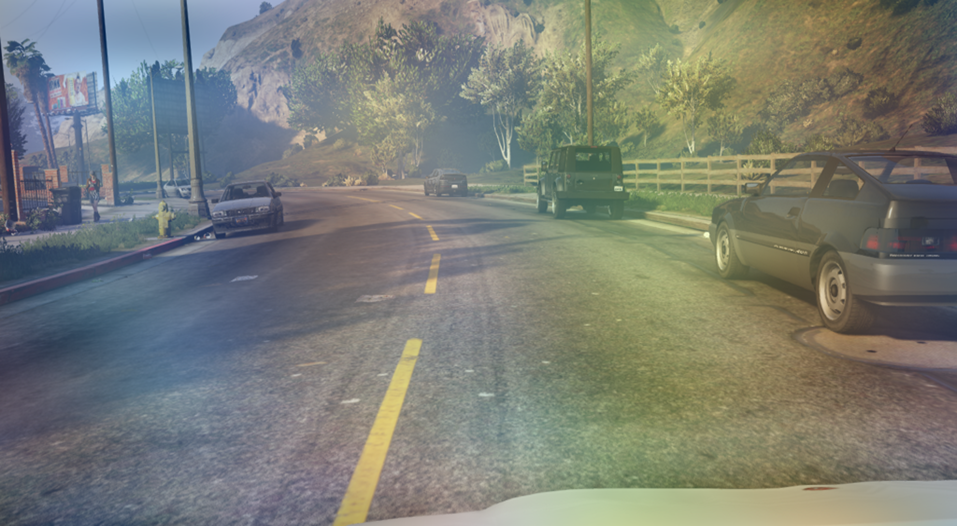}
    \end{subfigure}

    \caption{\textbf{Comparison to other image-to-image translation techniques.} Left column, top down: source image, MUINT, CUT, color transfer. Right column, top down: ProCST, BDL, CycleGAN, FDA.}
    \label{sit_compare}
\end{figure*}

We provide qualitative comparison between the translation of ProCST and this of 6 other image-to-image translation methods: BDL \cite{li2019bidirectional}, CUT \cite{park2020contrastive}, CycleGAN \cite{zhu2017unpaired}, MUINT \cite{huang2018munit}, color transfer \cite{reinhard2001color} and FDA \cite{yang2020fda}. We can see from Figure~\ref{sit_compare} that the first four methods hallucinate objects such as trees in the sky and Mercedes stars either on the road or on the mountains. Additionally, we can see that the translation of those methods distorts objects and creates blurry image. The final two methods do not hallucinate objects, but they fail to create a meaningful translation in terms of textures and colors. We can clearly see that the translation of ProCST outputs a clean and sharp image, that both preserve the semantic content of the source image and mimics the textures and colors of the target domain.

\subsection{Ablation Study - Qualitative Results}
We turn to discuss the differences and similarities between GTA5 $\rightarrow$ Cityscapes translated images generated from three different ProCST ablation models: full ProCST model, ProCST model trained with no label loss (i.e. $\lambda_{labels}=0$), and ProCST model trained with only one scale (i.e. $N=1$). The quantitative results of all three models appear in Section~\ref{sec:experiment}.

\begin{figure*}
    \begin{subfigure}{0.5\linewidth}
        \centering
        \includegraphics[width=\linewidth]{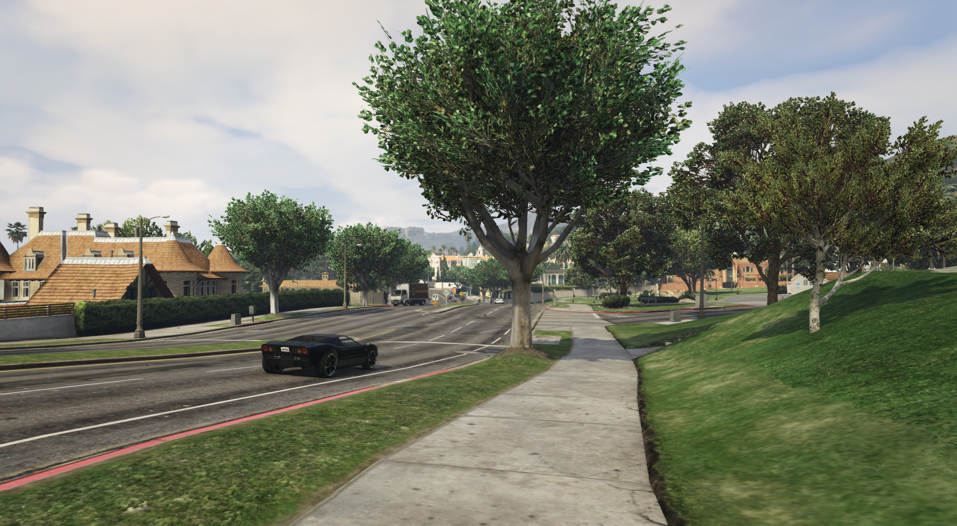}
    \end{subfigure}
    \begin{subfigure}{0.5\linewidth}
        \centering
        \includegraphics[width=\linewidth]{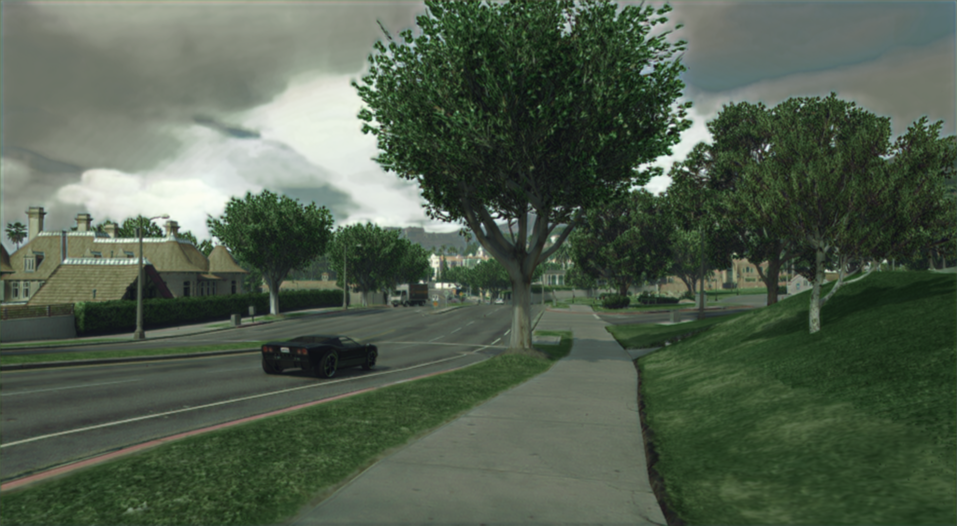}
    \end{subfigure}
        \begin{subfigure}{0.5\linewidth}
        \centering
        \includegraphics[width=\linewidth]{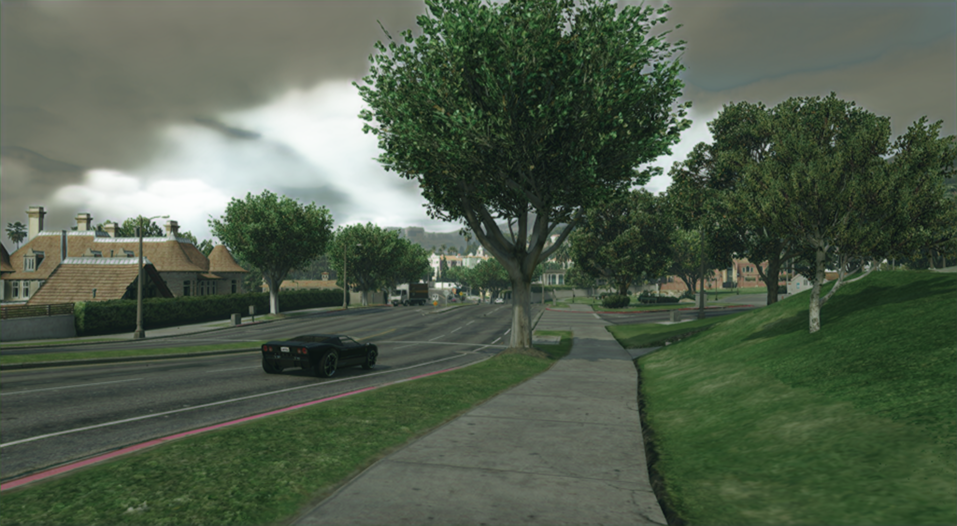}
    \end{subfigure}
    \begin{subfigure}{0.5\linewidth}
        \centering
        \includegraphics[width=\linewidth]{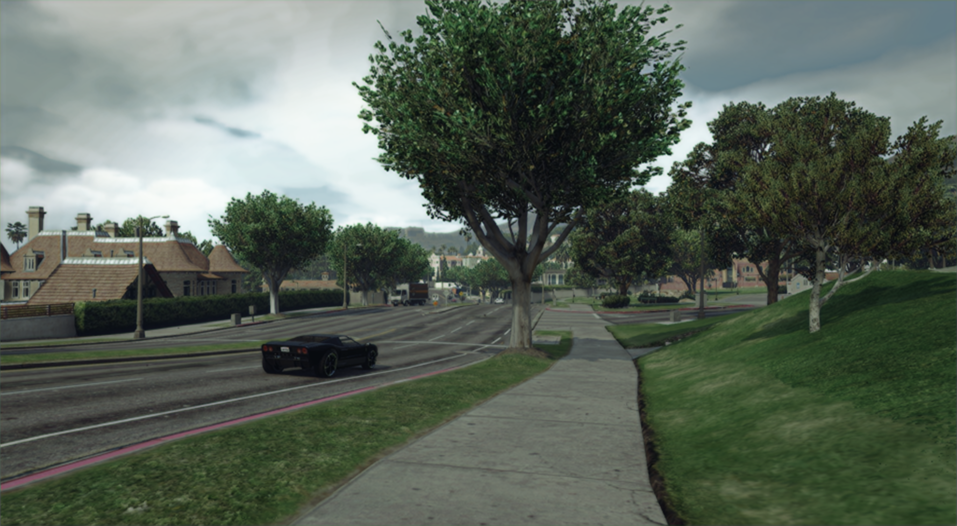}
    \end{subfigure}
    \caption{\textbf{Ablation Study.} Top Left: original image; Top Right: ProCST full model; Bottom Left: ProCST with $\lambda_{labels}=0$; Bottom Right: ProCST model $N=1$.}
    \label{ablations_compare_1}
\end{figure*}

\begin{figure*}
    \begin{subfigure}{0.5\linewidth}
        \centering
        \includegraphics[width=\linewidth]{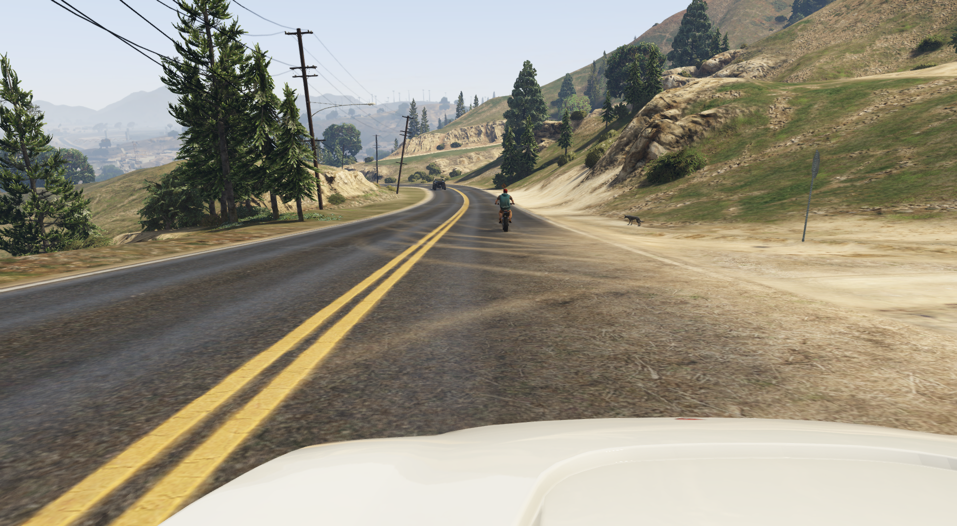}
    \end{subfigure}
    \begin{subfigure}{0.5\linewidth}
        \centering
        \includegraphics[width=\linewidth]{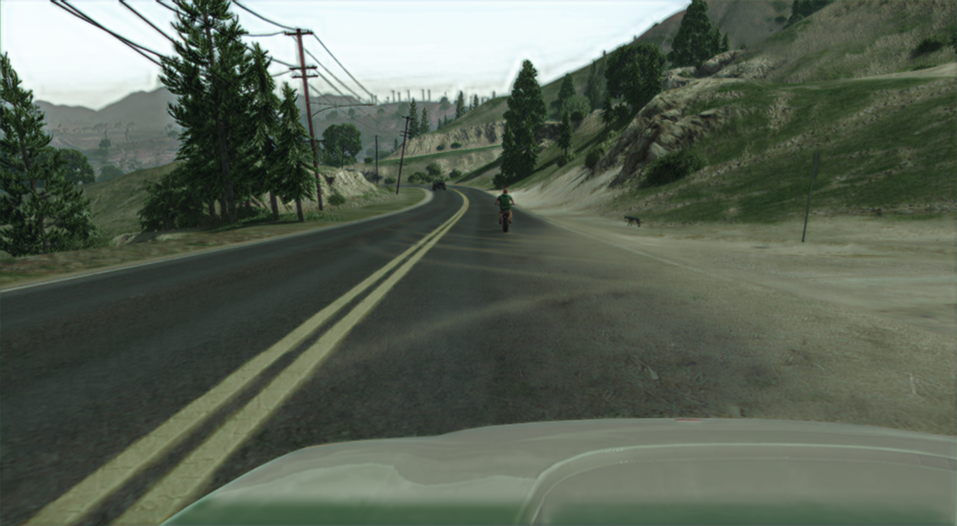}
    \end{subfigure}\\
    \begin{subfigure}{0.5\linewidth}
        \centering
        \includegraphics[width=\linewidth]{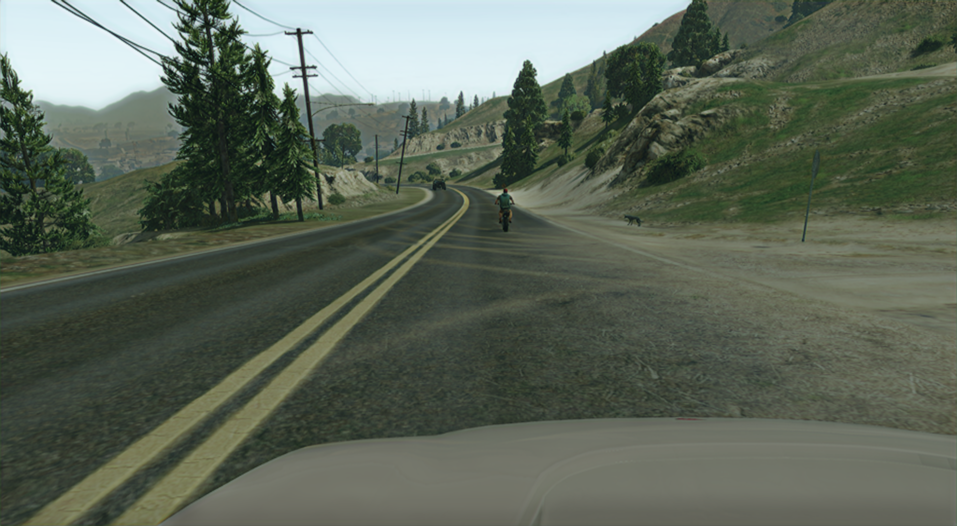}
    \end{subfigure}
    \begin{subfigure}{0.5\linewidth}
        \centering
        \includegraphics[width=\linewidth]{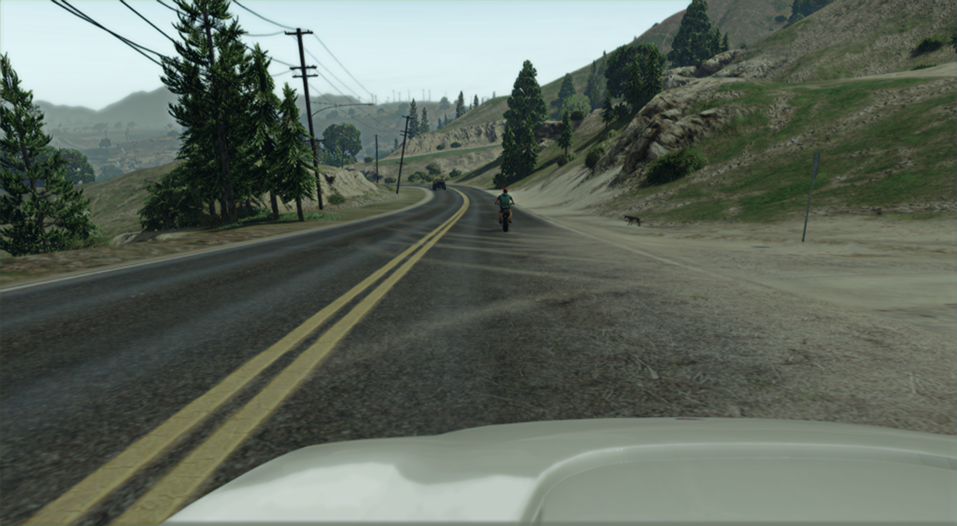}
    \end{subfigure}
    \caption{\textbf{Ablation Study.} Top Left: original image; Top Right: ProCST full model; Bottom Left: ProCST with $\lambda_{labels}=0$; Bottom Right: ProCST with $N=1$.}
    \label{ablations_compare_2}
\end{figure*}

\begin{figure*}
    \begin{subfigure}{0.5\linewidth}
        \centering
        \includegraphics[width=\linewidth]{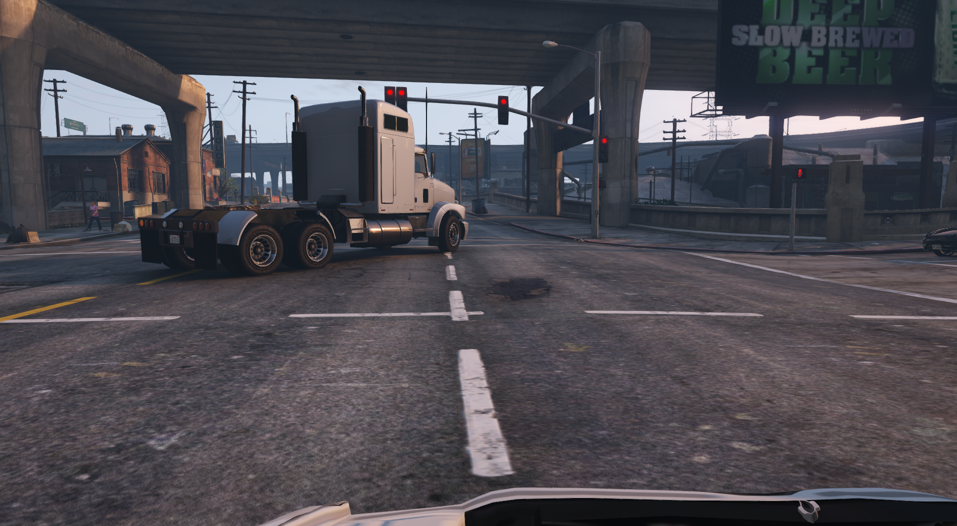}
    \end{subfigure}
    \begin{subfigure}{0.5\linewidth}
        \centering
        \includegraphics[width=\linewidth]{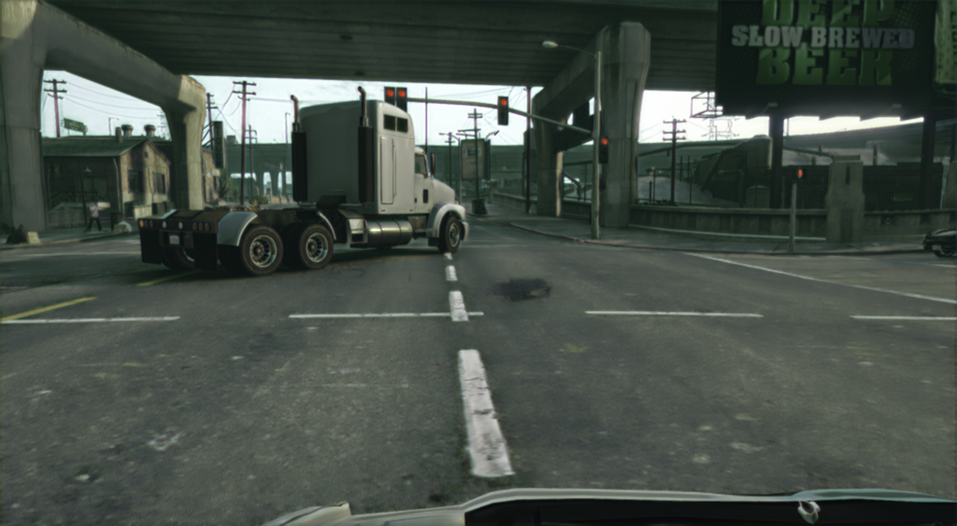}
    \end{subfigure}
        \begin{subfigure}{0.5\linewidth}
        \centering
        \includegraphics[width=\linewidth]{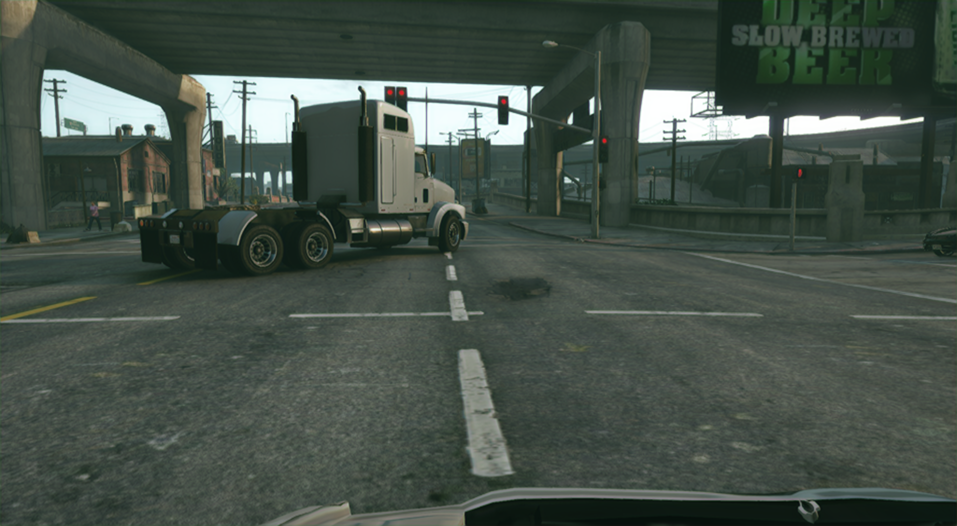}
    \end{subfigure}
    \begin{subfigure}{0.5\linewidth}
        \centering
        \includegraphics[width=\linewidth]{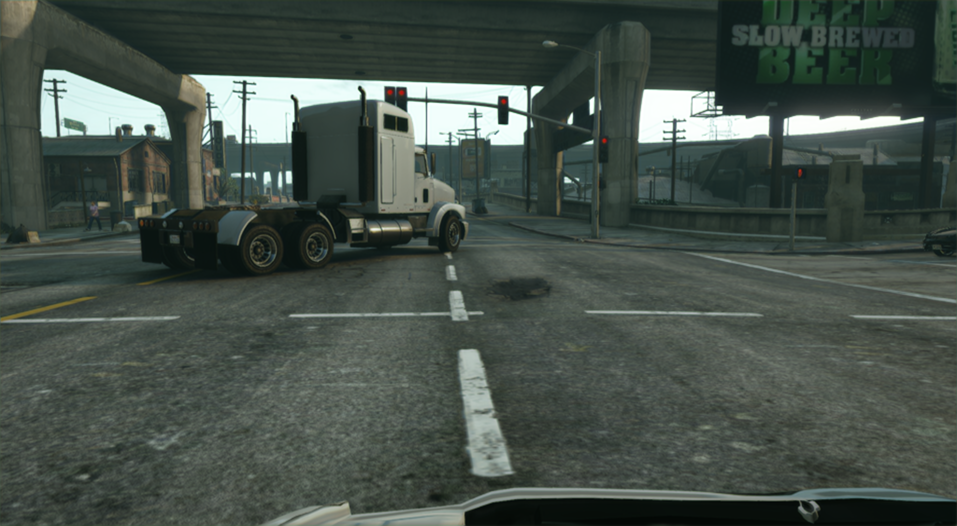}
    \end{subfigure}
    \caption{\textbf{Ablation Study.} Top Left: original image; Top Right: ProCST full model; Bottom Left: ProCST with $\lambda_{labels}=0$; Bottom Right: ProCST with $N=1$.}
    \label{ablations_compare_3}
\end{figure*}

\begin{figure*}
    \begin{subfigure}{0.5\linewidth}
        \centering
        \includegraphics[width=\linewidth]{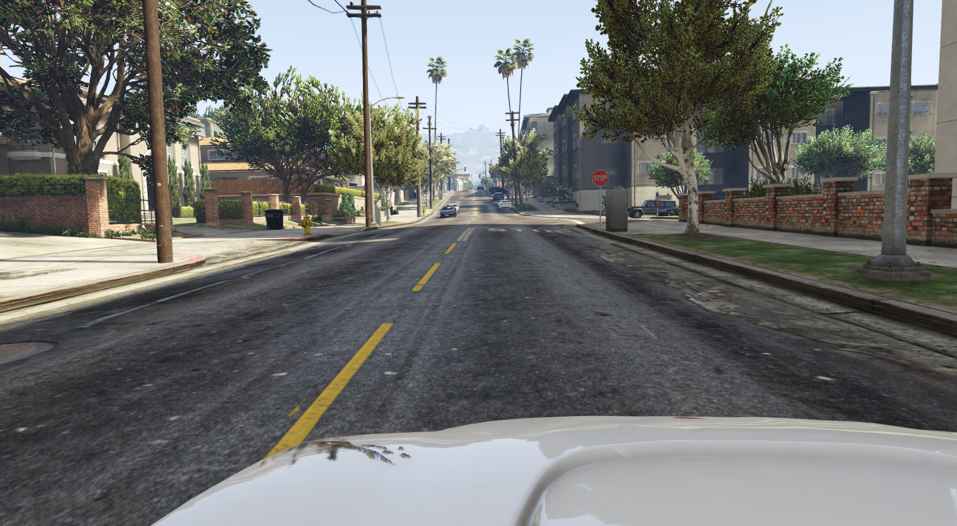}
    \end{subfigure}
    \begin{subfigure}{0.5\linewidth}
        \centering
        \includegraphics[width=\linewidth]{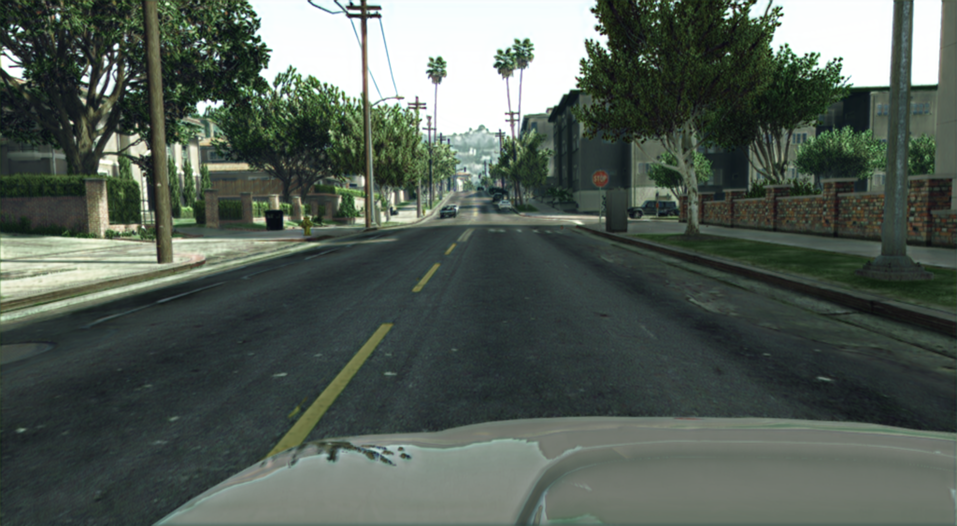}
    \end{subfigure}
    \begin{subfigure}{0.5\linewidth}
        \centering
        \includegraphics[width=\linewidth]{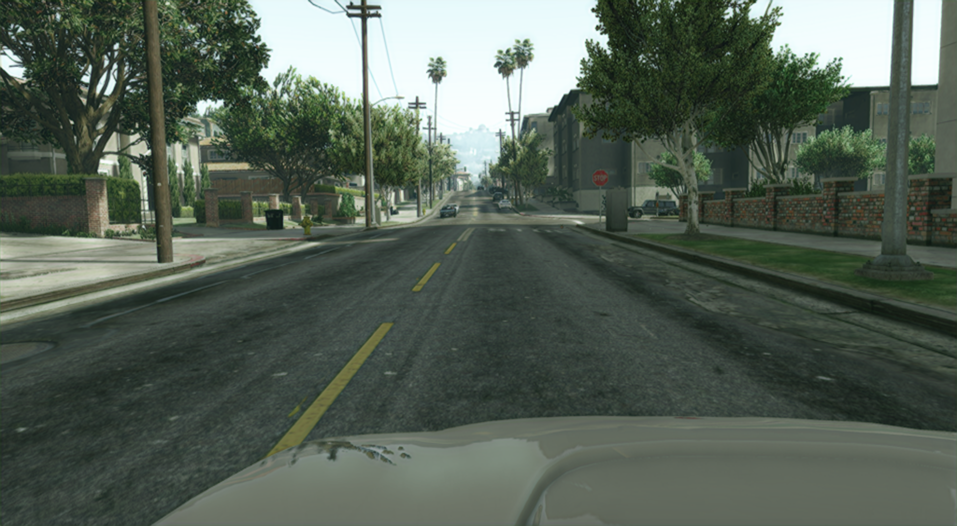}
    \end{subfigure}
    \begin{subfigure}{0.5\linewidth}
        \centering
        \includegraphics[width=\linewidth]{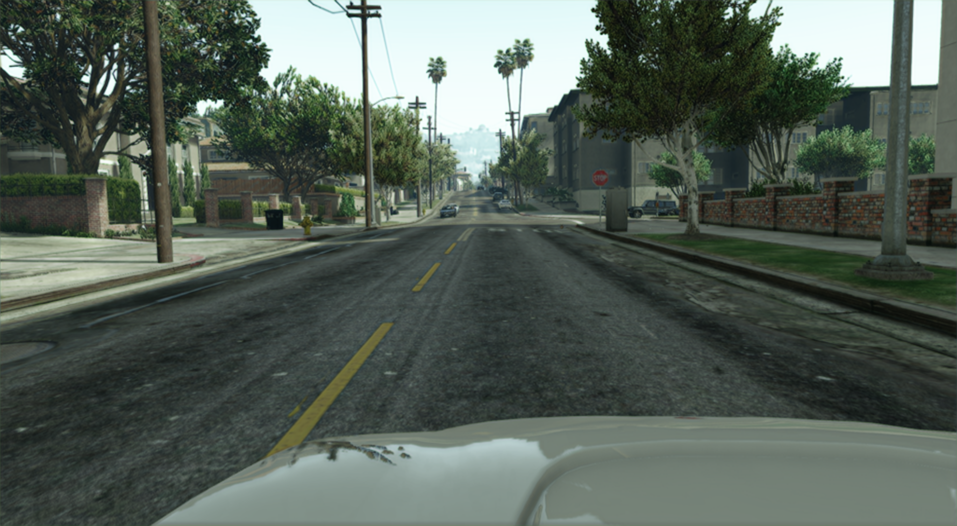}
    \end{subfigure}
    \caption{\textbf{Ablation Study.} Top Left: original image; Top Right: ProCST full model; Bottom Left: ProCST with $\lambda_{labels}=0$; Bottom Right: ProCST with $N=1$.}
    \label{ablations_compare_4}
\end{figure*}

Figures \ref{ablations_compare_1}, \ref{ablations_compare_2}, \ref{ablations_compare_3} and \ref{ablations_compare_4} present the original GTA5 image and all other SiT images generated using the compared models. Clearly, all three models create cloudy and green atmosphere as can be seen from Figure~\ref{ablations_compare_1}. This property matches Cityscapes domain.

Yet, in figures \ref{ablations_compare_2}, \ref{ablations_compare_3} and  \ref{ablations_compare_4} we can see visually that the full ProCST model adapts to the Cityscapes domain better than all other model. Visual quality, textures and general appearance are superior.  

\begin{figure*}
    \centering
    \begin{subfigure}{0.49\linewidth}
        \centering
        \includegraphics[width=\linewidth]{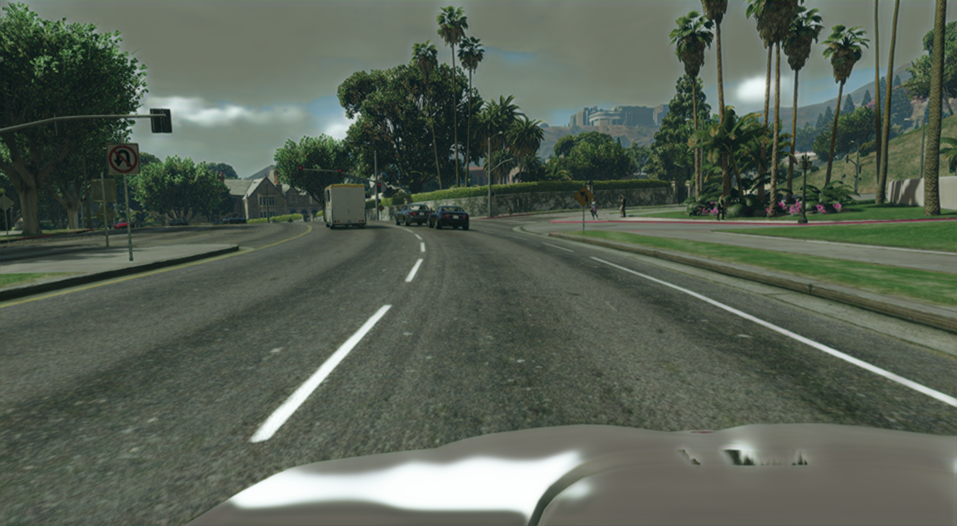}
    \end{subfigure}
    \begin{subfigure}{0.49\linewidth}
        \centering
        \includegraphics[width=\linewidth]{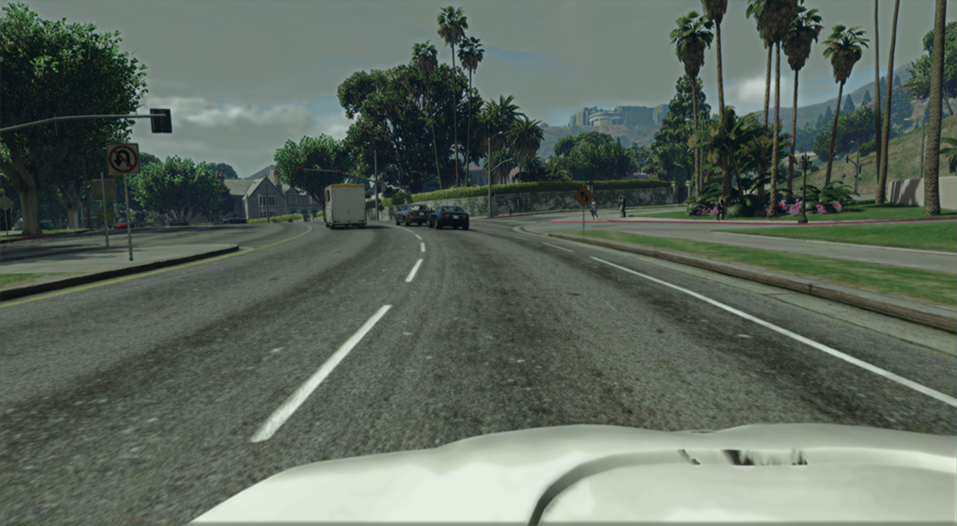}
    \end{subfigure}
    
    \begin{subfigure}{0.49\linewidth}
        \centering
        \includegraphics[width=\linewidth]{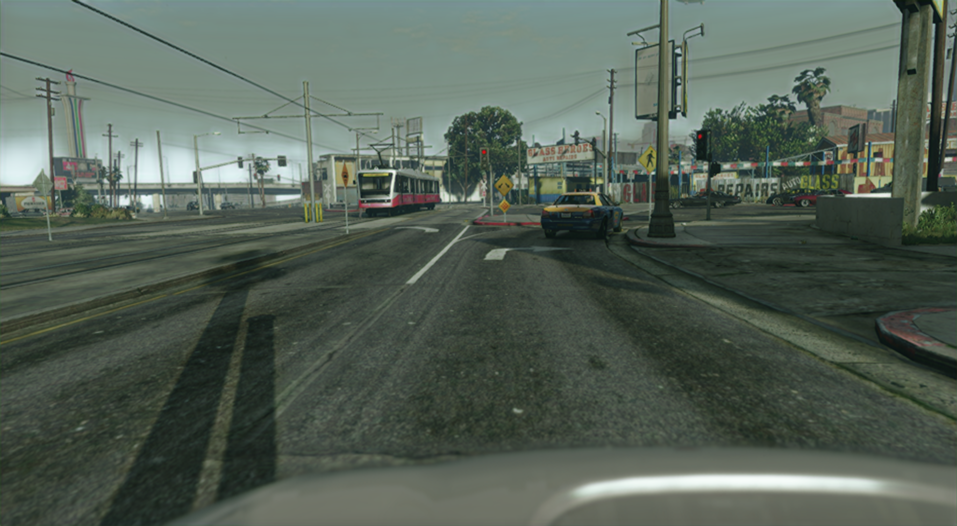}
    \end{subfigure}
    \begin{subfigure}{0.49\linewidth}
        \centering
        \includegraphics[width=\linewidth]{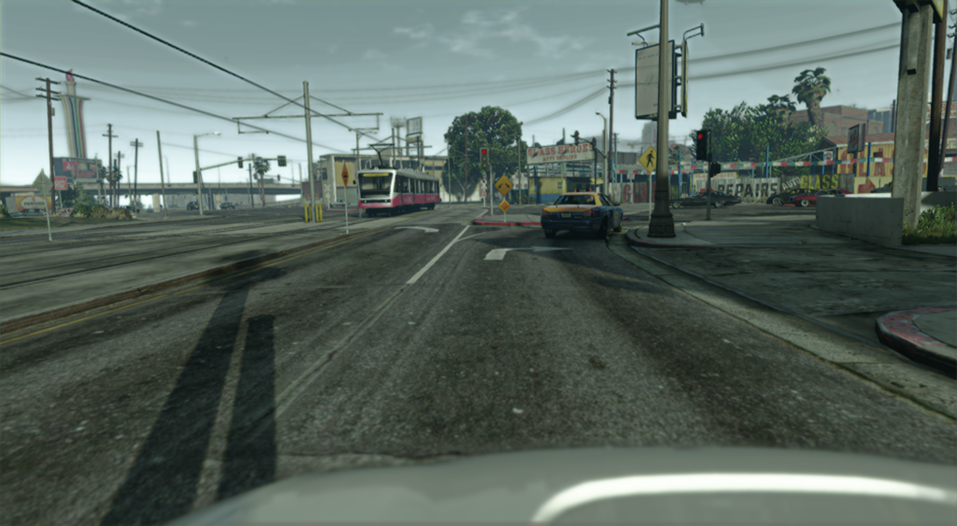}
    \end{subfigure}
    \caption{\textbf{Ablation Study.} Left: ProCST with $\lambda_{labels}=0$; Right: ProCST with $N=1$.}
    \label{no_seg_vs_no_ms}
\end{figure*}

Images \ref{ablations_compare_3} and \ref{ablations_compare_4} show that the ablated models fail to generate good SiT images when dealing with complicated texture translations, like the smoothness of the road.
Moreover, we can see that the quality of both ablated models is consistent with the quantitative results presented in the paper. While both variants are comparable in the results, we can see some minor differences between the ProCST model trained without the label loss term (i.e. $\lambda_{labels}=0$) and the one trained without multiscale structure (i.e. $N=1$). Multiscale structure helps to translate complicated textures, as we can see for example from the road's texture. Road is smoother and with deeper colors in the model with $\lambda_{labels}=0$ compared to the one with $N=1$. Despite this fact, the first model sometimes suffers from artifacts that the latter does not. We can see such a trend in Figure~\ref{no_seg_vs_no_ms}. In the first row, observe the smoothness of the road and the deep green color that has superior quality in $\lambda_{labels}=0$ than $N=1$. Yet, in the second row, the sky of the $N=1$ model is natural and clean while the sky of $\lambda_{labels}=0$ suffers from minor artifacts.

\subsection{SiT and TiS Translation Examples}
\begin{figure*}
\centering
    \begin{subfigure}{0.24\linewidth}
        \centering
        \includegraphics[width=\linewidth]{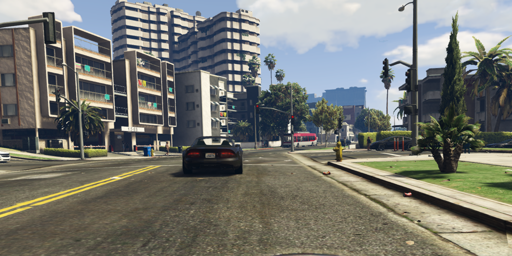}
    \end{subfigure}
    \begin{subfigure}{0.24\linewidth}
        \centering
        \includegraphics[width=\linewidth]{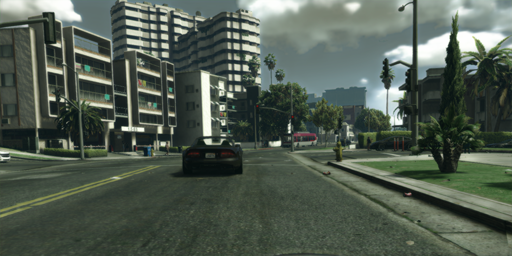}
    \end{subfigure}
    \begin{subfigure}{0.24\linewidth}
        \centering
        \includegraphics[width=\linewidth]{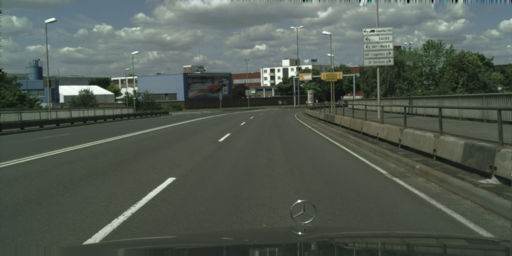}
    \end{subfigure}
    \begin{subfigure}{0.24\linewidth}
        \centering
        \includegraphics[width=\linewidth]{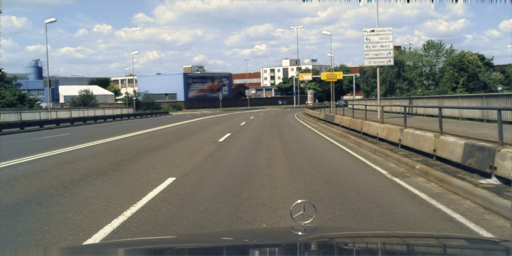}
    \end{subfigure}
    
    \begin{subfigure}{0.24\linewidth}
        \centering
        \includegraphics[width=\linewidth]{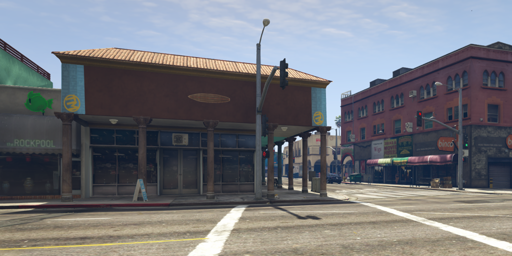}
    \end{subfigure}
    \begin{subfigure}{0.24\linewidth}
        \centering
        \includegraphics[width=\linewidth]{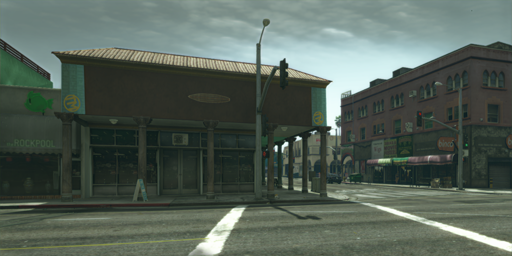}
    \end{subfigure}
    \begin{subfigure}{0.24\linewidth}
        \centering
        \includegraphics[width=\linewidth]{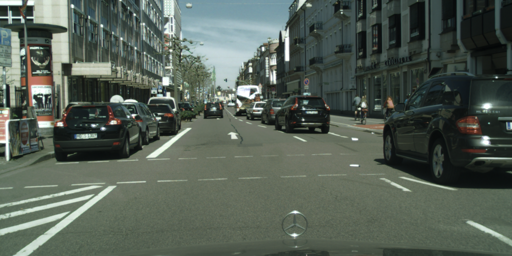}
    \end{subfigure}
    \begin{subfigure}{0.24\linewidth}
        \centering
        \includegraphics[width=\linewidth]{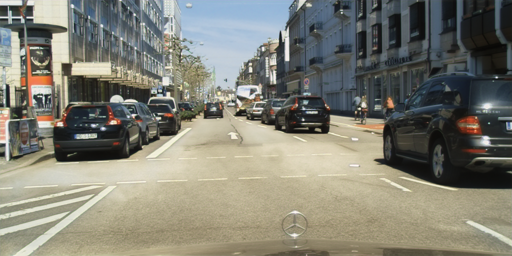}
    \end{subfigure}
    
    \begin{subfigure}{0.24\linewidth}
        \centering
        \includegraphics[width=\linewidth]{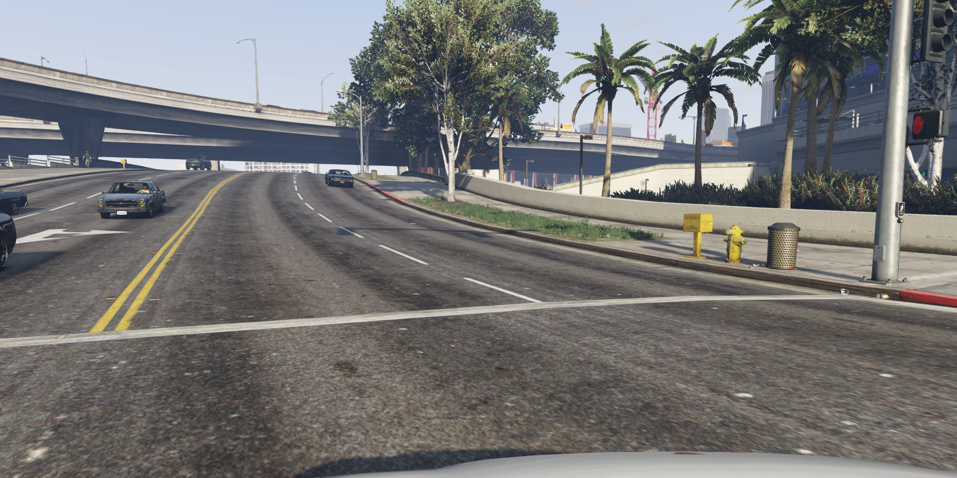}
    \end{subfigure}
    \begin{subfigure}{0.24\linewidth}
        \centering
        \includegraphics[width=\linewidth]{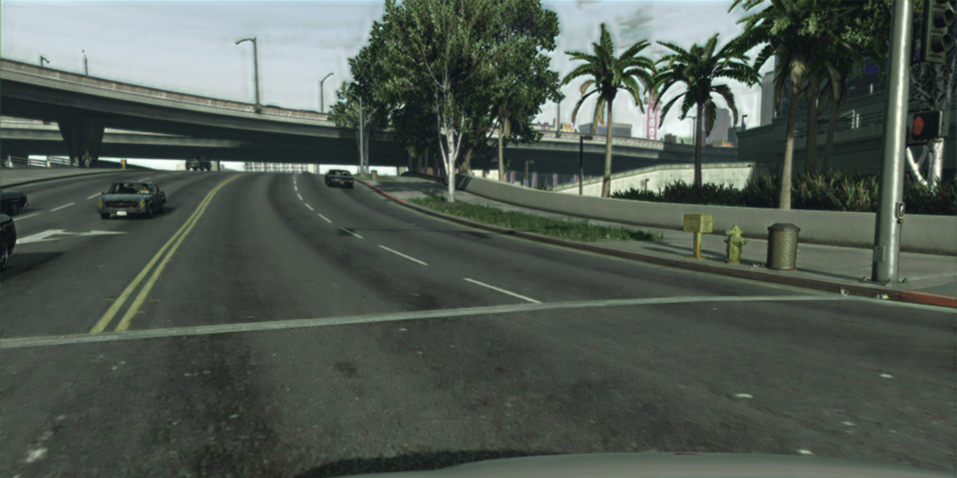}
    \end{subfigure}
    \begin{subfigure}{0.24\linewidth}
        \centering
        \includegraphics[width=\linewidth]{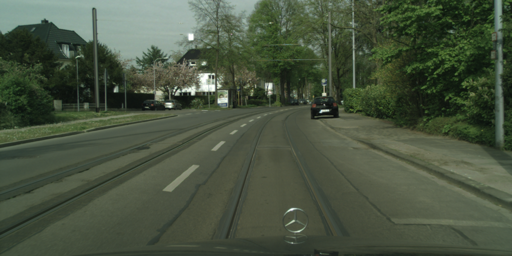}
    \end{subfigure}
    \begin{subfigure}{0.24\linewidth}
        \centering
        \includegraphics[width=\linewidth]{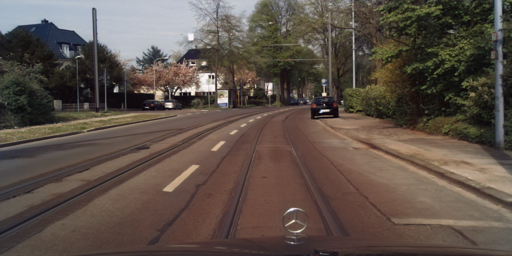}
    \end{subfigure}
    
    \begin{subfigure}{0.24\linewidth}
        \centering
        \includegraphics[width=\linewidth]{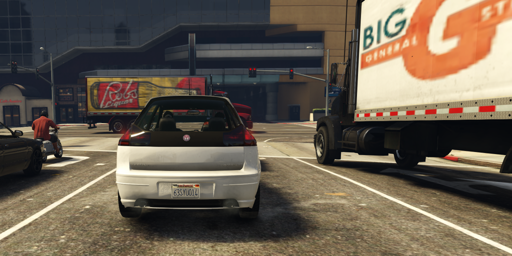}
    \end{subfigure}
    \begin{subfigure}{0.24\linewidth}
        \centering
        \includegraphics[width=\linewidth]{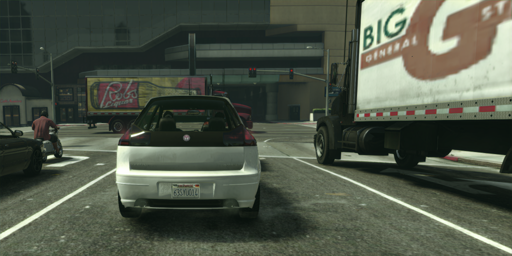}
    \end{subfigure}
    \begin{subfigure}{0.24\linewidth}
        \centering
        \includegraphics[width=\linewidth]{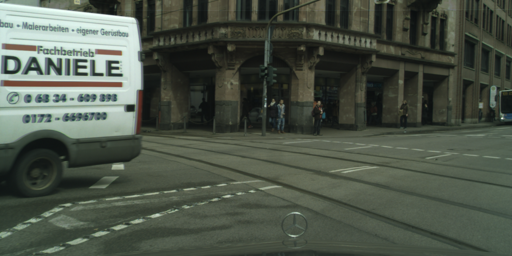}
    \end{subfigure}
    \begin{subfigure}{0.24\linewidth}
        \centering
        \includegraphics[width=\linewidth]{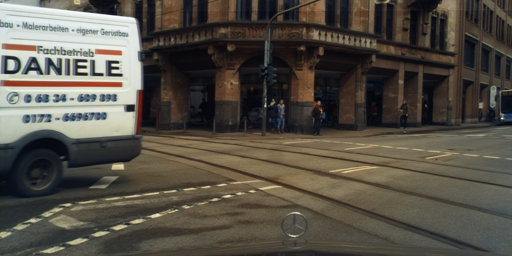}
    \end{subfigure}
    
    \begin{subfigure}{0.24\linewidth}
        \centering
        \includegraphics[width=\linewidth]{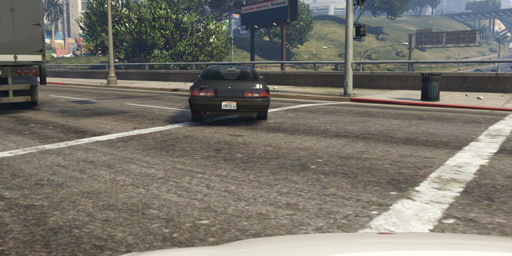}
    \end{subfigure}
    \begin{subfigure}{0.24\linewidth}
        \centering
        \includegraphics[width=\linewidth]{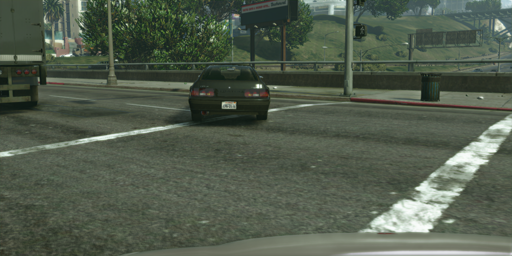}
    \end{subfigure}
    \begin{subfigure}{0.24\linewidth}
        \centering
        \includegraphics[width=\linewidth]{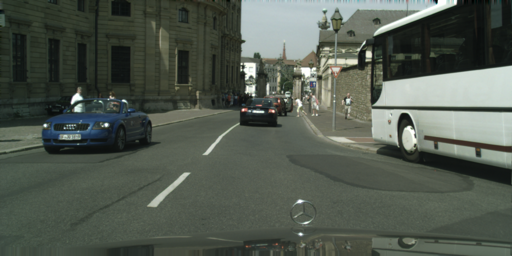}
    \end{subfigure}
    \begin{subfigure}{0.24\linewidth}
        \centering
        \includegraphics[width=\linewidth]{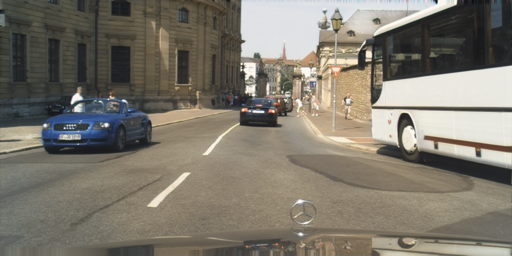}
    \end{subfigure}
    
    
    \begin{subfigure}{0.24\linewidth}
        \centering
        \includegraphics[width=\linewidth]{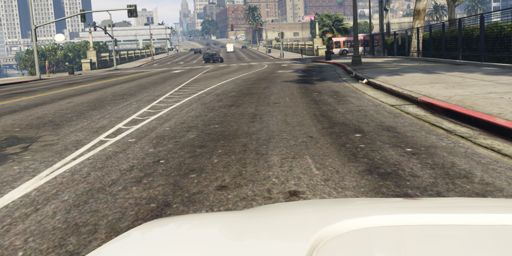}
    \end{subfigure}
    \begin{subfigure}{0.24\linewidth}
        \centering
        \includegraphics[width=\linewidth]{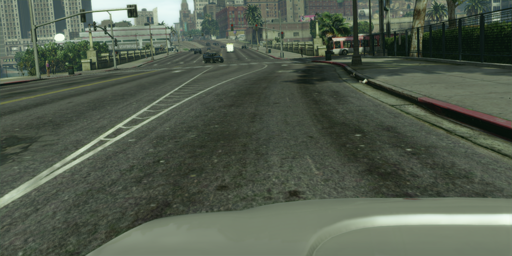}
    \end{subfigure}
    \begin{subfigure}{0.24\linewidth}
        \centering
        \includegraphics[width=\linewidth]{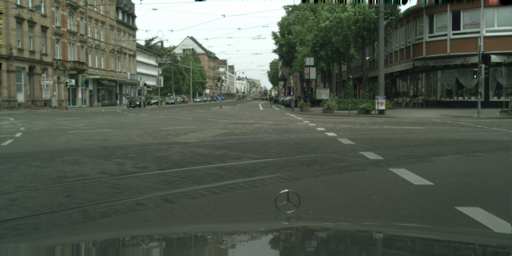}
    \end{subfigure}
    \begin{subfigure}{0.24\linewidth}
        \centering
        \includegraphics[width=\linewidth]{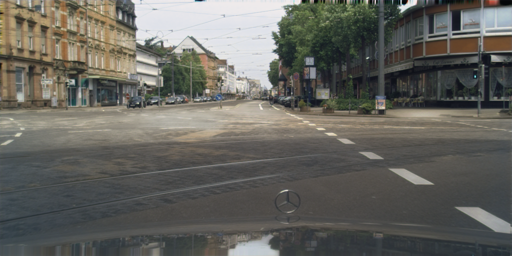}
    \end{subfigure}
    \caption{\textbf{Bi-directional $\text{GTA5} \xleftrightarrow[]{} \text{Cityscapes}$ image translation.} Left to right: GTA5, SiT, Cityscapes, and TiS.}
    \label{fig:sit_tis_gta_translations}
\end{figure*}

\begin{figure*}
    \centering
    \begin{subfigure}{0.24\linewidth}
        \centering
        \includegraphics[width=\linewidth]{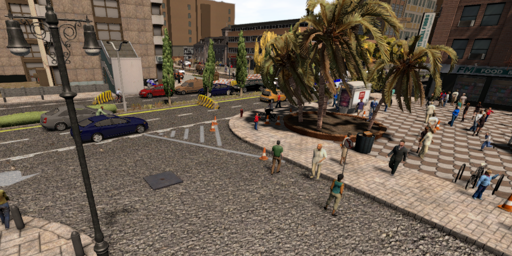}
    \end{subfigure}
    \begin{subfigure}{0.24\linewidth}
        \centering
        \includegraphics[width=\linewidth]{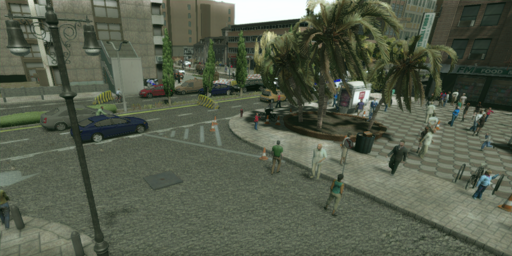}
    \end{subfigure}
    \begin{subfigure}{0.24\linewidth}
        \centering
        \includegraphics[width=\linewidth]{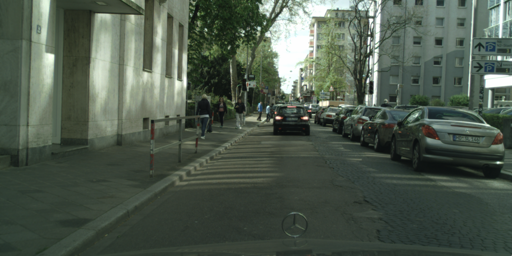}
    \end{subfigure}
    \begin{subfigure}{0.24\linewidth}
        \centering
        \includegraphics[width=\linewidth]{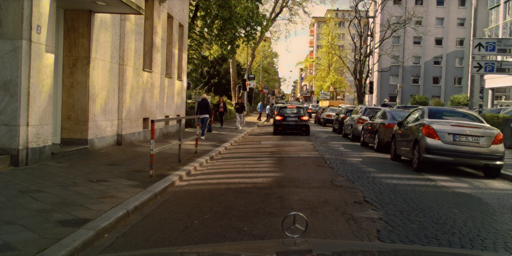}
    \end{subfigure}
    
    \centering
    \begin{subfigure}{0.24\linewidth}
        \centering
        \includegraphics[width=\linewidth]{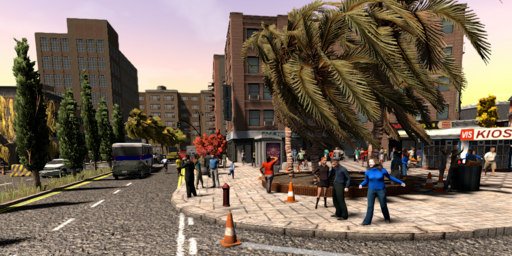}
    \end{subfigure}
    \begin{subfigure}{0.24\linewidth}
        \centering
        \includegraphics[width=\linewidth]{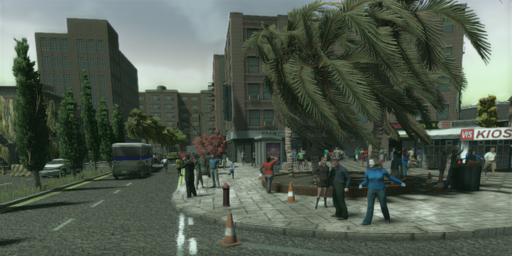}
    \end{subfigure}
    \begin{subfigure}{0.24\linewidth}
        \centering
        \includegraphics[width=\linewidth]{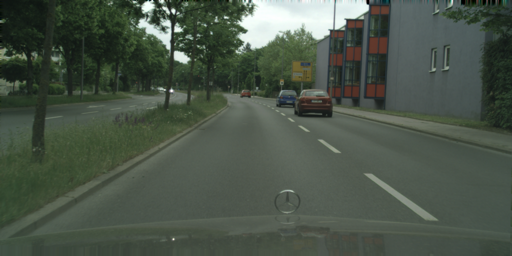}
    \end{subfigure}
    \begin{subfigure}{0.24\linewidth}
        \centering
        \includegraphics[width=\linewidth]{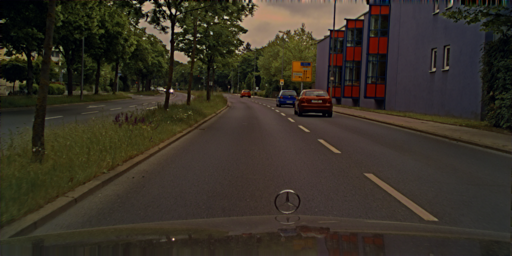}
    \end{subfigure}

    \centering
    \begin{subfigure}{0.24\linewidth}
        \centering
        \includegraphics[width=\linewidth]{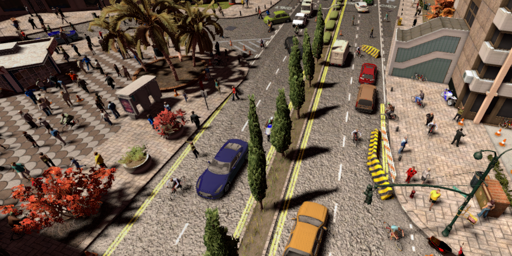}
    \end{subfigure}
    \begin{subfigure}{0.24\linewidth}
        \centering
        \includegraphics[width=\linewidth]{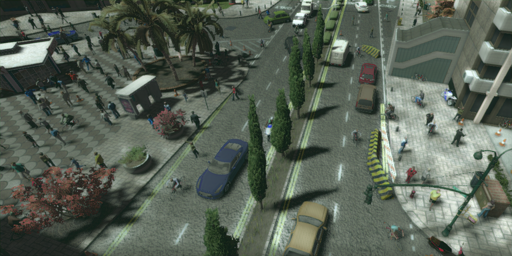}
    \end{subfigure}
        \begin{subfigure}{0.24\linewidth}
        \centering
        \includegraphics[width=\linewidth]{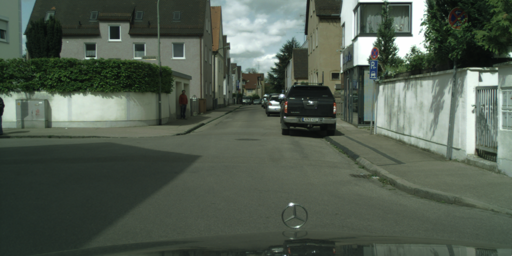}
    \end{subfigure}
    \begin{subfigure}{0.24\linewidth}
        \centering
        \includegraphics[width=\linewidth]{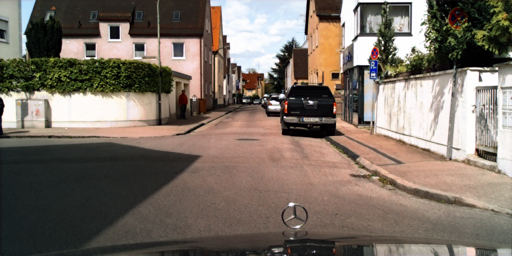}
    \end{subfigure}
    
    \centering
    \begin{subfigure}{0.24\linewidth}
        \centering
        \includegraphics[width=\linewidth]{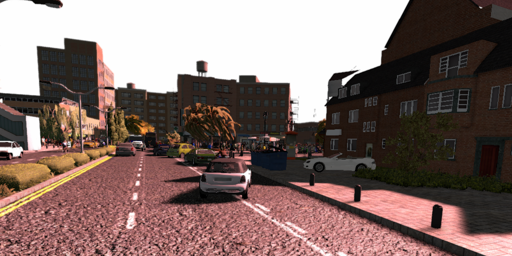}
    \end{subfigure}
    \begin{subfigure}{0.24\linewidth}
        \centering
        \includegraphics[width=\linewidth]{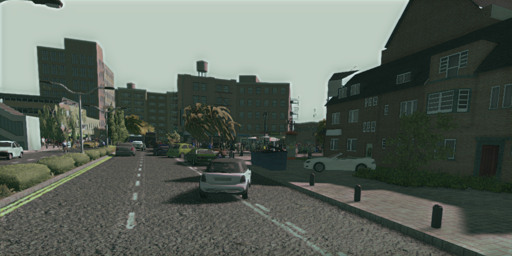}
    \end{subfigure}
    \begin{subfigure}{0.24\linewidth}
        \centering
        \includegraphics[width=\linewidth]{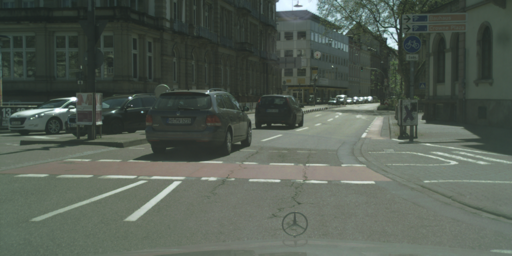}
    \end{subfigure}
    \begin{subfigure}{0.24\linewidth}
        \centering
        \includegraphics[width=\linewidth]{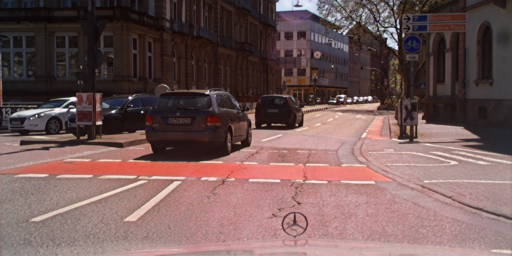}
    \end{subfigure}
    \begin{subfigure}{0.24\linewidth}
        \centering
        \includegraphics[width=\linewidth]{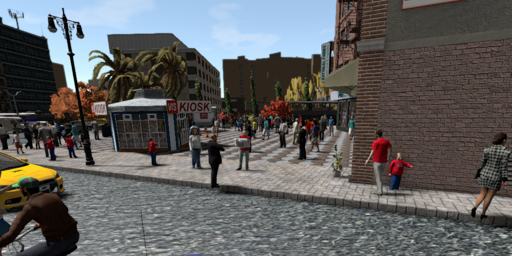}
    \end{subfigure}
    \begin{subfigure}{0.24\linewidth}
        \centering
        \includegraphics[width=\linewidth]{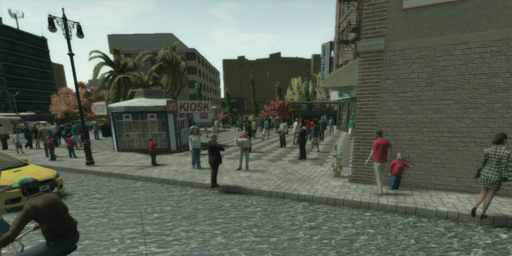}
    \end{subfigure}
    \begin{subfigure}{0.24\linewidth}
        \centering
        \includegraphics[width=\linewidth]{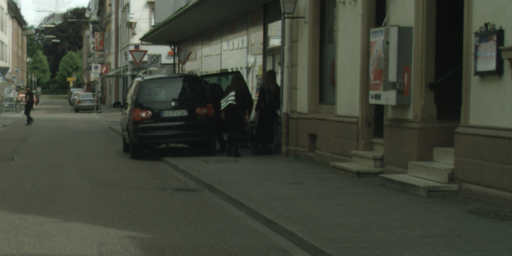}
    \end{subfigure}
    \begin{subfigure}{0.24\linewidth}
        \centering
        \includegraphics[width=\linewidth]{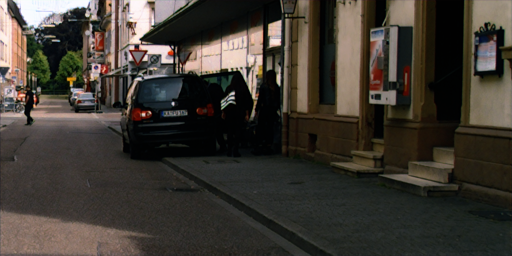}
    \end{subfigure}
    \begin{subfigure}{0.24\linewidth}
        \centering
        \includegraphics[width=\linewidth]{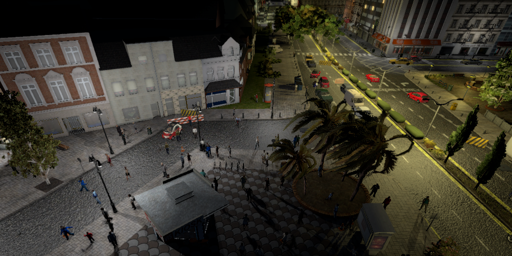}
    \end{subfigure}
    \begin{subfigure}{0.24\linewidth}
        \centering
        \includegraphics[width=\linewidth]{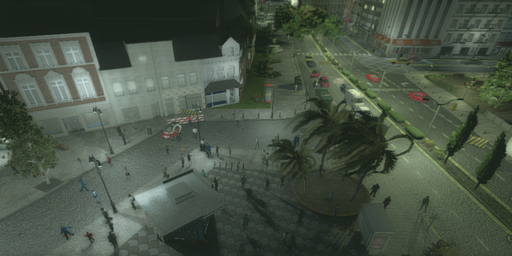}
    \end{subfigure}
    \begin{subfigure}{0.24\linewidth}
        \centering
        \includegraphics[width=\linewidth]{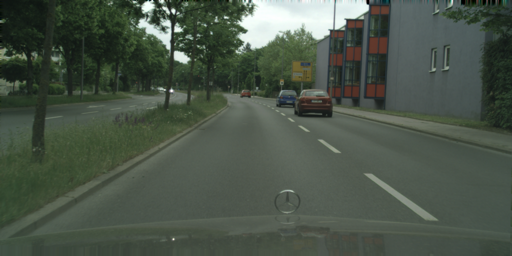}
    \end{subfigure}
    \begin{subfigure}{0.24\linewidth}
        \centering
        \includegraphics[width=\linewidth]{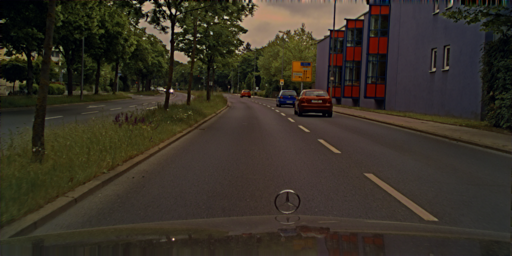}
    \end{subfigure}
    \caption{\textbf{Bi-directional $\text{Synthia} \xleftrightarrow[]{} \text{Cityscapes}$ image translation.} Left to right: Synthia, SiT, Cityscapes, and TiS images.}
    \label{fig:sit_tis_synthia_translations}
\end{figure*}

We provide additional examples of SiT and TiS images side by side with the original source images, for both source domains. Figure~\ref{fig:sit_tis_gta_translations} shows GTA5 $\xleftrightarrow{}$ Cityscapes translations and Figure~\ref{fig:sit_tis_synthia_translations} shows Synthia $\xleftrightarrow{}$ Cityscapes translations.


\clearpage
\bibliography{aaai23}

\end{document}